\documentclass{article}

\usepackage{arxiv}

\usepackage{amssymb}

\usepackage{amsmath}
\usepackage[utf8]{inputenc} 
\usepackage[T1]{fontenc}    
\usepackage{hyperref}       
\usepackage{url}            
\usepackage{booktabs}       
\usepackage{amsfonts}       
\usepackage{nicefrac}       
\usepackage{microtype}      
\usepackage{lipsum}		
\usepackage{graphicx}
\usepackage{natbib}
\usepackage{doi}

\usepackage{url}
\usepackage{mathrsfs}  
\usepackage{enumitem}
\usepackage{siunitx}
\usepackage{tikz}
\usetikzlibrary{fit,positioning}
\usepackage{algorithm}
\usepackage{algpseudocode}
\algnewcommand{\LineComment}[1]{\State \(\triangleright\) #1}

\DeclareMathOperator*{\argmax}{arg\,max}

\usepackage{subfigure}
\usepackage{caption}

\usepackage{cleveref}

\title{On the use of feature-maps and parameter control for improved quality-diversity meta-evolution}
\author{David M. Bossens\\
	University of Southampton\\
	\texttt{D.M.Bossens@soton.ac.uk} \\
	\And
	Danesh Tarapore \\
	University of Southampton\\
}

\begin{document}

\maketitle

\begin{abstract}
In Quality-Diversity (QD) algorithms, which evolve a behaviourally diverse archive of high-performing solutions, the behaviour space is a difficult design choice that should be tailored to the target application. In QD meta-evolution, one evolves a population of QD algorithms to optimise the behaviour space based on an archive-level objective, the meta-fitness. This paper proposes an improved meta-evolution system such that (i) the database used to rapidly populate new archives is reformulated to prevent loss of quality-diversity; (ii) the linear transformation of base-features is generalised to a feature-map, a function of the base-features parametrised by the meta-genotype; and (iii) the mutation rate of the QD algorithm and the number of generations per meta-generation are controlled dynamically. Experiments on an 8-joint planar robot arm compare feature-maps (linear, non-linear, and feature-selection), parameter control strategies (static, endogenous, reinforcement learning, and annealing), and traditional MAP-Elites variants, for a total of 49 experimental conditions. Results reveal that non-linear and feature-selection feature-maps yield a 15-fold and 3-fold improvement in meta-fitness, respectively, over linear feature-maps. Reinforcement learning ranks among top parameter control methods. Finally, our approach allows the robot arm to recover a reach of over 80\% for most damages and at least 60\% for severe damages.
\end{abstract}


{\let\thefootnote\relax\footnotetext{This work is the extended version of the paper:\\ \textit{David M. Bossens \& Danesh Tarapore (2021). On the use of feature-maps for
improved quality-diversity meta-evolution. In 2021 Genetic and Evolutionary
Computation Conference Companion (GECCO '21 Companion), July 10--14,
2021, Lille, France. ACM, New York, NY, USA, 2 pages. https://doi.org/10.1145/3449726.3459442}}} 

\section{Introduction}
Historically, most evolutionary algorithms (EAs) were designed to optimise a fitness function, solving a single problem without considerations for generalisation to unseen problems or robustness to perturbations to the evaluation environment. However, it was widely known that successfully converging to the maximum of that fitness function requires maintaining genetic diversity in the population of solutions (see e.g., \cite{Laumanns2002,Gupta2012,Ursem2002,Ginley2011}). Moreover, the use of niching demonstrated how  maintaining subpopulations could help find multiple solutions to a single problem \citep{Mahfoud1995}. Some studies included genetic diversity as one of the objectives of the EA \citep{Toffolo2003}. Approaches in evolutionary robotics, artificial life, and neuro-evolution realised that genetic diversity does not necessarily imply a diversity of solutions, since (i) different genotypes may encode the same behaviour and vice versa (especially for complex genotypes such as neural networks); and (ii) many genotypes may encode unsafe or undesirable solutions that should be discarded during evolution (e.g., self-collisions on a multi-joint robot arm). Such approaches began to emphasise \textit{behavioural diversity} \citep{Mouret2009a,Gomez2009,Mouret2009,Mouret2012a}, not only as a driver for objective-based evolution but also as the enabler for diversity- or novelty-driven evolution \citep{Lehman2011}.

In \textit{quality-diversity algorithms} such as MAP-Elites  
\citep{Mouret2015} and Novelty Search with Local Competition \citep{LehmanStanley2011}, the behavioural diversity approach is combined with local competition such that for each local region in the behaviour space the best solution is stored, forming a large archive of solutions -- a \textit{behaviour-performance map} in the case of MAP-Elites. The development of quality-diversity algorithms has allowed a plethora of useful applications. In robotics, this includes the design of robot morphologies and controllers \citep{Mouret2015,NordmoenEllefsen2018} and behaviour adaptation \citep{Cully2015b}, in which a robot recovers from environmental changes or damages to its sensory-motor system by searching for high-performing controllers across the evolved archive of solutions.

Two important design choices of a quality-diversity algorithm are its behaviour space, the behavioural features that define the behavioural diversity across the solutions, as well as its various hyperparameters, such as the mutation rate, the population size, or the crossover or mutation operator. Traditionally, the behavioural features are chosen by the user to fit the purpose. However, complex features that are non-intuitive to the user may optimise the intended purpose better. Similarly, the best evolutionary parameters to achieve the highest quality-diversity metrics are typically unknown. Therefore, an automated approach to the behaviour space and evolutionary parameters may be required.

This paper explores an approach to \textit{quality-diversity meta-evolution}, in which the behaviour space and evolutionary parameters are automatically determined to optimise a meta-objective. We explore (i) how to evolve a low-dimensional behaviour space by means of a feature-map, a function taking as input the ``meta-genotype'' and a larger number behavioural base-features and outputting the target-features, and (ii) control strategies to automatically determine the mutation rate and the number of generations (per meta-generation) of the quality-diversity algorithm represented by the meta-genotype.  As an application, we consider an 8-joint planar robotic arm where the meta-objective is adaptation to damages. The meta-objective is to be achieved by a combination of diversity, defined as a wide variety of robot poses, and quality, defined as robust and efficient movements characterised by low variance.
\section{Related work}
Behaviour adaptation to unforeseen damages is one of the main applications of our work. Prior work has shown that performing a search across the behaviour-performance maps evolved by MAP-Elites can give high-performing recovery solutions within a limited number of function evaluations \citep{Cully2015b}. We are interested in how automating the behaviour space and the evolutionary hyperparameters affects the adaptation.

In automating the behaviour space, there are three competing classes of methods. The first class formulates the behaviour space by means of unsupervised learning, forming a generative model of behaviour \citep{Nguyen2016,Cully2019,Cully2018a}, for example,  based on an auto-encoder neural network that represents a compressed and low-dimensional model of sensory data \citep{Cully2019,Cully2018a}. Such approaches can reduce dimensionality and form a robust denoised behaviour space. The second class models the genotypes of the elites (i.e., the highest-quality solutions) and biases the search to find such elites (e.g., by adapting the crossover operator based on a generative model of the behaviour-performance map) \citep{Gaier2020}. This class allows to rapidly find high-performing solutions according to the fitness function with a high coverage of solutions. Both classes do not address how the behaviour space is to be selected to maximise a custom archive-level objective. The third class, of which our method is a member, formulates the behaviour space to maximise an archive-level objective within a particular domain of problems. \cite{Meyerson2016} proposed to learn a behavioural distance function for Novelty Search to suit a particular domain of problems. In Meyerson et al. (2016), the weights given to a large number of base-features depended heuristically on the behaviours that were successful on the target domain. \cite{Bossens2020a} proposed a quality-diversity meta-evolution system, where one also adapts weights given to a large number of base-features but where the weights are evolved by an evolutionary algorithm to define a lower-dimensional behaviour space for MAP-Elites. We expand on this work by (i) reformulating the database to prevent the loss of quality-diversity, (ii) generalising the linear combination of base-features to a feature-map, a parametrised function of the base-features, and (iii) considering a more complete meta-evolution system that further optimises the MAP-Elites algorithm by dynamically controlling its various hyperparameters.

Our study investigates the effects of various parameter control strategies. Although this topic has been explored widely (for an overview, see  \cite{Eiben2003,Karafotias2015,Doerr19tutorial}), it is rarely explored in quality-diversity algorithms. To the best of our knowledge, one prior work has explored parameter control of the mutation rate in quality-diversity algorithms \citep{Nordmoen2018}. We investigate the mutation rate as well as the generations per meta-generation within a quality-diversity meta-evolution system.\\
\section{Quality-diversity meta-evolution with feature-maps and parameter control}
MAP-Elites (ME) evolves a behaviour-performance map, storing the highest-fitness controllers for each hypercube in a discretised behaviour space \citep{Mouret2015}. Since ME is not explicitly optimised for generalisation, we use Meta-evolution with CMA-ES \citep{Bossens2020a} to evolve a population of MEs with generalisation as a meta-objective. In Meta-evolution with CMA-ES, low-dimensional behaviour spaces were automatically generated from a weighted sum of a higher-dimensional base-behavioural space. Due to storing all solutions generated by prior MAP-Elites algorithms in a database, Meta-evolution with CMA-ES allowed to efficiently populate behaviour-performance maps on-the-fly, without the need to evolve solutions from scratch. To construct an improved implementation, we propose three modifications to Meta-evolution with CMA-ES:
\begin{itemize}[noitemsep,nolistsep]
\item a novel type of database that prevents the loss of behaviourally diverse and high-performing solutions (see Section~\ref{subsec: database});
\item a more generic feature-map that allows non-linear transformations of the base-features (see Section~\ref{subsec: featuremaps});
\item the use of parameter control, or dynamic optimisation, of evolutionary hyperparameters, namely the number of generations per meta-generation and the mutation rate (see Section~\ref{subsec: meta-optimisation}).
\end{itemize} %
\subsection{MAP-Elites algorithm}
The MAP-Elites algorithm discretises the behaviour space into behavioural bins, which are equally-sized hypercubes, and then maintains for each behavioural bin the elite solution (i.e., the solution with the highest fitness), leading to quality-diversity.

MAP-Elites first randomly generates an initial population of genotypes. Then, each genotype in the initial population is evaluated, resulting in a fitness score $f$ and a behavioural descriptor $\mathbf{\beta}$. Each genotype is then added to the behaviour-performance map $\mathcal{M}$ based on the following replacement rule: if the behavioural bin for $\mathbf{\beta}$ is empty (i.e., $\mathcal{M}[\mathbf{\beta}]=\emptyset$) or if the fitness is higher than the current genotype in that bin (i.e., $ f > f(\mathcal{M}[\mathbf{\beta}])$), then place the genotype $\mathbf{g}$ in that bin of the behaviour-performance map (i.e., $\mathcal{M}[\mathbf{\beta}] \gets \mathbf{g}$).

After initialisation, the algorithm applies repeated cycles of random selection, genetic variation, evaluation, and replacement. Random selection is implemented by randomly selecting genotypes from non-empty behavioural bins in the behaviour-performance maps. Genetic variation is based on mutations to the genotypes. Evaluation of genotypes is based on a user-defined fitness function $f(\cdot)$. Replacement is based on the above-mentioned replacement rule. After many repetitions of this cycle, the behaviour-performance map is gradually filled with a behaviourally diverse and high-quality solutions.

\subsection{Meta-evolution with CMA-ES}
\label{subsec: meta-optimisation}
The behaviour space in MAP-Elites is not necessarily optimised for generalisation. To automatically improve the behaviour space towards a generalisation-based meta-objective, we use the Covariance Matrix Adaptation Evolutionary Strategy (CMA-ES; \cite{Hansen2007,Hansen2016}), following in this regard the Meta-evolution with CMA-ES \citep{Bossens2020a}. 

The algorithm first applies an initialisation phase to populate the behaviour-performance maps at the first meta-generation. A large number of random genotypes are sampled, evaluated, and then added to the database $\mathcal{D}$ (see Section~\ref{subsec: database} for details).

After initialisation, the meta-evolution algorithm performs a large number of \textit{meta-generations}, consisting of four main steps.

In the first step of the meta-generation, the meta-generation constructs new maps $\mathcal{M}_i$, for each $i \in \{1,\dots,\lambda\}$ in the meta-population (see l.~8-14 in Algorithm~\ref{alg: meta-CMAES}), based on the meta-genotype and the existing solutions in the database. After initialising an empty map, new meta-genotypes $\mathbf{w} \in \mathbb{R}^n$ are sampled based on a multivariate normal distribution, 
\begin{equation}
\label{eq: multivariate normal}
\mathbf{w} \sim \mathcal{N} \! \left( \mathbf{m}, \sigma \mathbf{C} \right) \,,
\end{equation}
where $\mathbf{m} \in \mathbb{R}^n$ is the mean meta-genotype, $\mathbf{C}$ is the covariance matrix, and $\sigma > 0$ is a scalar representing the step-size. Each meta-genotype $\mathbf{w} \in \mathbb{R}^n$ is then transformed from a vector into a more useful format, denoted by $\mathbf{W}$ -- for example, for the linear feature-map in Eq.~\ref{eq: linfm} $\mathbf{W}$ is a matrix, whereas for the non-linear feature-map in Eq.~\ref{eq: linfm} $\mathbf{W}$ consists of two weight matrices and two biases. Each entry in the database $\langle \mathbf{g}, \mathbf{b}, f  \rangle \in \mathcal{D}$ is then processed to obtain the resulting behavioural descriptor $\mathbf{\beta}$ via the feature-map in Eq.~\ref{eq: featuremap}, and if MAP-Elites' replacement rule is satisfied then the entry is added to the map according to $\mathcal{M}[\mathbf{\beta}] \gets \mathbf{g}$. 

\sloppy In the second step of the meta-generation,, after all behaviour-performance maps are filled with database entries,\footnote{Empirical tests show populating one behaviour-performance map from the large database of around 5 million solutions consumes on average \SI{6}{s}.} the meta-individuals $i \in \{1,\dots,\lambda \}$ independently apply MAP-Elites, evolving their own behaviour-performance map $\mathcal{M}_i$ for a number of $I$ iterations. During these iterations of MAP-Elites (see l.~30-38  in Algorithm~\ref{alg: meta-CMAES}), new bottom-level genotypes are formed by mutating existing bottom-level genotypes in the map. Each created solution is given a behavioural description according to the feature-map $\mathbf{\beta} \gets \phi(\mathbf{W}, \mathbf{b})$ (see Section~\ref{subsec: featuremaps}),  put in the behaviour-performance map if the replacement rule is satisfied, and added to the database (see l.~23-29 in Algorithm~\ref{alg: meta-CMAES}).  

In the third step, each meta-individual is evaluated on the meta-fitness $\mathcal{F}(\mathbf{w})$, which represents a map-level objective such as the adaptation performance of the behaviour-performance map in various unseen contexts (see Section \ref{sec: experimental-setup} for its implementation). 

In the final step of the meta-generation, CMA-ES updates the mean, covariance and step size parameters (see l.~19-21 in Algorithm~\ref{alg: meta-CMAES}), applying the $(\mu/\mu_W, \lambda)$-CMA Evolution Strategy \citep{Hansen2016} to optimise the meta-fitness. One selects the $\mu \leq \lambda$ individuals with higest meta-fitness for reproduction. Reproduction involves mutating the mean towards the best of the selected meta-individuals,
\begin{equation}
\label{eq: update-mean}
\mathbf{m} \gets \mathbf{m} + c_m \sigma \sum_{i=1}^{\mu} v_i (\mathbf{w}^{i} - \mathbf{m}) \,,
\end{equation}
where $v_i > 0$, $\sum_{i=1}^{\mu} v_i = 1$; $\mathbf{w}^ {i}$ is the $i$'th best meta-genotype; $\sigma$ is the step size; and $c_m \in [0,1]$ is a learning rate. The covariance matrix is adapted based on a combination of the active rank-$\mu$ update \citep{Jastrebski2006}, which exploits information from the entire population by assigning positive weights to highest-ranking individuals and negative weights to lowest-ranking individuals, and the rank-one update \citep{Hansen1996}, which exploits the correlations between generations based on the evolution path,
\begin{equation}
\label{eq: update-covariance}
\mathbf{C} \gets \left(1 - c_1 - c_{\mu} \sum_j v_j\right) \mathbf{C}   + c_{\mu} \sum_{i=1}^{\lambda} v_i \mathbf{s}_i \mathbf{s}_i^{\intercal} + c_1 \mathbf{p}_c \mathbf{p}_c^{\intercal}  \,,
\end{equation} 
where $c_{\mu}$ and $c_1$ are positive weights reflecting the importance of the rank-$\mu$ and rank-one term, respectively; $\mathbf{s}_i \sim \mathcal{N}(\mathbf{0},\mathbf{C})$ is the difference of the sampled meta-genotype from the old mean, divided by the step size $\sigma$; $v_i$ is a positive scalar in case $i \leq \mu$ and a negative scalar otherwise ; and $\mathbf{p}_c \in \mathbb{R}^n$ is the evolution path, a weighted sum of the past mutation steps. Finally, the step-size is controlled with cumulative step-size control,
\begin{equation}
\label{eq: update-sigma}
\sigma \gets \sigma \exp \left( \frac{c_{\sigma}}{d_{\sigma}} \left( \frac{|| \mathbf{p}_{\sigma}  ||}{\mathbb{E}\left[|| \mathcal{N} \! \left( \mathbf{0}, \mathbf{I} \right) || \right]} - 1 \right) \right)    \,,
\end{equation}
where $c_{\sigma}$ and $d_{\sigma}$ are parameters that affect the damping and $\mathbf{p}_{\sigma} \in \mathbb{R}^n$ is the conjugate evolution path -- which is a weighted sum of the past mutation steps but unlike $\mathbf{p}_c$ it is formulated such that its expected Euclidian norm is independent of its direction. When the conjugate evolution path is longer than expected, the successive steps were positively correlated, and the step size will be increased to reduce the number of steps to reach a promising region in search space. When the conjugate evolution path is shorter than expected, successive steps were negatively correlated, and the step size will be decreased to avoid successive steps cancelling out each other.

\begin{algorithm}
  \caption{Meta-evolution with CMA-ES.}
  \label{alg: meta-CMAES}
    \begin{algorithmic}[1] 
      \State $\mathcal{D} \gets \emptyset$. \Comment{Create empty database.}
      \For{$i=1$ to $p$} \Comment{Create initial database.}
      \State $\mathbf{g} \gets \texttt{random-genotype()}$. 
      \State $\mathbf{b}, f \gets \texttt{eval}(\mathbf{g})$.  \Comment{Base-features and fitness.}
      \State Insert $\langle \mathbf{g}, \mathbf{b}, f \rangle$ into $\mathcal{D}$. \Comment{Fill the database (see Section~\ref{subsec: database}).}
      \EndFor

      \For{$j=1$ to $G$ } \Comment{Loop over meta-generations.}
      	
      	\For {$i=1$ to $\lambda$}
      		\State Set $\mathcal{M}^i \gets \emptyset$. \Comment{Empty the map.}
      		\State $\mathbf{w} \sim \mathcal{N} \! \left( \mathbf{m}, \sigma \mathbf{C} \right)$. \Comment{Sample meta-genotype.}
      		\For { $\langle \mathbf{g},\mathbf{b},f \rangle \in \mathcal{D}$ } \Comment{Construct map from database.}
      		\State \texttt{add-to-map}($\mathcal{M}^i$, $\mathbf{w}$, $\mathbf{g}$, $\mathbf{b}$, $f$).
      		\EndFor
      	\EndFor
      	\For {$i=1$ to $\lambda$}
      		\State Perform \texttt{MAP-Elites-iterations}($\mathcal{M}^i$,$\mathbf{w}^i$).
      		\State $\mathcal{F}_i \gets$ \texttt{Meta-fitness}($\mathcal{M}^i$). 
      	\EndFor
      	
      	\State $\mathbf{m} \gets $ \texttt{Update-mean}(). \Comment{ See Eq.~\ref{eq: update-mean}.}
      	\State $\mathbf{C} \gets $ \texttt{Update-covariance}(). \Comment{ See Eq.~\ref{eq: update-covariance}.}
      	\State $\sigma   \gets $ \texttt{Update-step}().\Comment{ See Eq.~\ref{eq: update-sigma}.}
      \EndFor 
      \Procedure{add-to-map}{$\mathcal{M}$, $\mathbf{w}$, $\mathbf{g}$, $\mathbf{b}$, $f$}
      \State $\mathbf{W} \gets \texttt{transform}(\mathbf{w})$. \Comment{Convert to useful format (see Section~\ref{subsec: featuremaps}).}
      \State $\mathbf{\beta} \gets \phi(\mathbf{W},\mathbf{b})$. \Comment{Apply feature-map to get target features (see Eq.~\ref{eq: featuremap}-\ref{eq: nonlinfm}).}
      \If {$\mathcal{M}[\mathbf{\beta}] = \emptyset$ \textbf{ or } $f > f(\mathcal{M}[\mathbf{\beta}])$}
      \State $\mathcal{M}[\beta] \gets \mathbf{g}$. \Comment{Add genotype $\mathbf{g}$ to the map $\mathcal{M}$.}      
      \EndIf
      \EndProcedure
      \Procedure{MAP-Elites-iterations}{$\mathcal{M}$, $\mathbf{w}$}
      \For {$i=1$ to $I$} \Comment{$I$ is the number of iterations}
      	\State $\mathbf{g} \sim \mathcal{M}$. \Comment{Sample genotype randomly from map.}
      	\State $\mathbf{g}' \gets \texttt{mutate}(\mathbf{g})$. \Comment{Mutation.}
      \State $\mathbf{b}, f \gets \texttt{eval}(\mathbf{g}')$.  \Comment{Base-features and fitness.}
	  \State \texttt{add-to-map}($\mathcal{M}$, $\mathbf{w}$, $\mathbf{g}'$, $\mathbf{b}$, $f$).
      \State Insert $\langle \mathbf{g}', \mathbf{b}, f \rangle$ into $\mathcal{D}$. \Comment{Fill the database (see Section~\ref{subsec: database}).}
      \EndFor
      \EndProcedure
    \end{algorithmic}
\end{algorithm}

\subsection{$k$-best database}
\label{subsec: database}
To enable rapidly generating new behaviour-performance maps, the database $\mathscr{D}$ stores a large number of previously found solutions. Each such solution is a tuple $\langle \mathbf{g}, \mathbf{b}, f  \rangle$, where $\mathbf{g}$ is the bottom-level genotype (e.g., the parameters of a controller), $\mathbf{b}$ is an extended behavioural description of the solution according to a large number of $N_b$ user-defined behavioural base-features, and $f$ is the fitness (e.g., the performance of a controller). To retain behavioural diversity, the database $\mathscr{D}$ divides the base-behavioural space $[0,1]^{N_b}$ into coarse-grained bins of equal width $\delta$, in which it stores up to a number of $k$ solutions. The hypercube partitioning is geometrically similar to the behaviour-performance maps in MAP-Elites except that (a) only a small number, such as 2 or 3, bins per dimension (corresponding to a width of $\delta=1/2$ or $\delta=1/3$) are allowed to limit the maximal capacity of the database to $2^{N_b}$ or $3^{N_b}$ given the many base-features; and (b) for each hypercube, an array containing up to at most $k$ solutions is stored within a single bin.

When a new solution $\langle \mathbf{g}, \mathbf{b}, f  \rangle$ is presented to the database, its corresponding coarse-grained bin is looked up, yielding the array of solutions $\mathscr{C} = \mathscr{D}[\mathbf{b}]$. Then an additional check for fitness and diversity is performed: if there is another solution in $\mathscr{C}$ that is in the same base-behavioural hypercube of width $\delta / k$, then the solution is only added if it has higher fitness; if there is no such similar solution, then the solution is always added.\footnote{Since there are $k^{N_b}$ possible hypercubes of width $\delta / k$ within each coarse-grained bin of width $\delta$ and up to $k$ solutions are allowed per coarse-grained bin, the check for diversity and fitness is not too restrictive.} If $\mathscr{C}$ now has $k+1$ solutions, then its lowest-fitness solution is removed. If the database's capacity is exceeded, the number of allowed solutions per bins is decremented, $k \gets k -1$, and for each coarse-grained bin  the lowest-fitness solution is removed. While $k$ is initially large, e.g., $k=1000$, $k$ shrinks progressively as the run continues.

\subsection{Feature-maps}
\label{subsec: featuremaps}
To transform the base-behavioural features $[0,1]^{N_b}$ to a low-dimensional behavioural descriptor $\mathbf{\beta} \in [0,1]^D$, we use a feature-map $\phi$, 
\begin{equation}
\label{eq: featuremap}
\mathbf{\beta} = \phi(\mathbf{W}, \mathbf{b}) \,,
\end{equation}
which is parametrised by the meta-genotype $\mathbf{W}$. 

We explore three instantiations of the feature-map. A first instance is the linear transformation as replicated from \cite{Bossens2020a},
\begin{equation}
\label{eq: linfm}
\mathbf{\beta} = \mathbf{W} \mathbf{b} \,,
\end{equation}
 where $\mathbf{W} \in \mathbb{R}^{D \times N_b}$. To allow improved coverage of the behaviour space, we add an expanding normalisation $\beta_i \gets (\beta_i - m)/(M - m)$ for all $i \in \{1,\dots,D\}$, where $m$ and $M$ are the minimum and maximum target-feature values, respectively, estimated from empirical data (see Section \ref{sec: experimental-parameters}). A second included feature-map, chosen to be able to demonstrate whether or not it is necessary to combine features, is a feature selector according to 
\begin{equation}
\label{eq: selectionfm}
\begin{split}   
j^{i} & = \argmax_{j \in N_b} W_{ij} \\
\mathbf{\beta} & = b_{j^{1}},\dots, b_{j^{D}}\,,
\end{split}
\end{equation}
such that for each feature of the resulting descriptor, $i=1,\dots,D$, the base-feature $b_j$ with highest $\mathbf{W}_{ij}$ is chosen. A third included feature-map is a non-linear transformation using a neural network with a single hidden layer with sigmoid activations. It is chosen to demonstrate the need for non-trivial feature-maps, and because neural networks of this type can represent arbitrary mappings due to their representational capacity. The neural network is given by
\begin{equation}
\label{eq: nonlinfm}
\begin{split}
f(\mathbf{W}, \mathbf{b}) & = \mathbf{W}^{2} S_{N_b}(\mathbf{W}^{1}  \mathbf{b} + B^1) + B^2 \\
\mathbf{\beta} & = S_{N_h}(f(\mathbf{W}, \mathbf{b})) \,,
\end{split}
\end{equation}
 where $S_N(\mathbf{x}) = 1 / \left( 1 + \exp\left(-\alpha_{s} \mathbf{x}/(N+1)\right)\right)$ is an elementwise sigmoid function; $\alpha_s$ is an empirically defined 
 scaling factor; and now $\mathbf{W}$ is composed of a weight matrix from input to hidden layer, $\mathbf{W}^{1} \in \mathbb{R}^{N_h \times N_b}$, a weight matrix from hidden layer to output, $\mathbf{W}^{2} \in \mathbb{R}^{D \times N_h}$, and the corresponding bias units $B^1, B^2 \in \mathbb{R}$. For networks such as $f(\mathbf{W}, \mathbf{b})$, but also those replacing the activation function with the wider class of non-polynomial piecewise continuous functions, universal approximation theorems (see e.g., \cite{Leshno1993,Hornik1989a}) show that in principle all multi-variate functions over closed and bounded intervals can be represented to arbitrary precision, assuming a sufficient amount of neurons. Adding a sigmoid transformation as we have implemented is therefore not strictly necessary for representational capacity.\footnote{However, the representational capacity is preserved. Let $\{(\mathbf{b}_1,\mathbf{\beta}_1),\dots,(\mathbf{b}_n,\mathbf{\beta}_n)\}$ be a finite set of input-output pairs. Since the sigmoid function is monotonously increasing, for every $\mathbf{\beta}_i \in [0,1]^D$, $i \in \{1,\dots,n\}$, there exists a unique $\mathbf{\beta}_i' \in \mathbb{R}^D$ such that $\mathbf{\beta}_i'  = S^{-1}(\mathbf{\beta}_i)$ and $\mathbf{\beta}_i'  \neq S^{-1}(\mathbf{\beta})$ for all $\mathbf{\beta} \neq \mathbf{\beta}_i'$. Due to not having an activation function at the output units, the universal approximation theorems apply and the function that maps pairs $\{(\mathbf{b}_1,\mathbf{\beta}_1'),\dots,(\mathbf{b}_n,
\mathbf{\beta}_n')\}$ can therefore be represented with arbitrary precision. In practice, restricting $\mathbf{\beta}_i' \in 
[-5,5]^D$ for all $i \in \{1,\dots,n\}$ is sufficient to approximate any $\mathbf{\beta}_i \in [0,1]^D$ accurately.} 
 
We require that for a sizeable proportion of mappings, varying the base-behavioural features in $[0,1]^{N_b}$ ensures (i) diversity, such that the entire output range in $[0,1]^D$ can be reached with high statistical probability, and (ii) quality, such that each behavioural bin has a large enough number of solutions for local competition. If the frequency of certain behavioural bins is extremely low for some bins and extremely high for others, then there cannot be both quality and diversity. The output sigmoid activation in Eq.~\ref{eq: nonlinfm} accounts for the high frequencies of near-zero values of $f(\mathbf{W}, \mathbf{b})$ as it increases steeply for values close to zero and slowly for extreme values. This ensures that for a sizeable proportion of mappings, each bin in $[0,1]^D$ is frequently represented, leading to quality-diversity.

\subsection{Dynamic parameter control}
While Meta-evolution with CMA-ES optimises the behavioural features of MAP-Elites, there are a few parameter settings which are difficult to tune and possibly require dynamic changes throughout the evolutionary run. Therefore, an additional change is the dynamic control of parameter settings for the MAP-Elites meta-individuals. In particular, we aim to optimise MAP-Elites' mutation rate and the number of iterations using dynamic parameter control strategies. The following algorithms are included (more details in Section \ref{sec: experimental-setup}):
\begin{itemize}[noitemsep,nolistsep]
\item Endogenous: an additional gene in the meta-genotype encodes the parameter.
\item Annealing: the parameter is annealed linearly from its maximal to its minimal value depending on the number of function evaluations.
\item Reinforcement learning: a reinforcement learning agent optimises the parameter based on the progress of the algorithm.
\item Static: we try out a limited number of static parameter settings (i.e., traditional tuning).
\end{itemize}
\section{Experiment setup}
\label{sec: experimental-setup}
\subsection{Simulation environment}
We use DART (Dynamic Animation and Robotics Toolkit; \cite{Lee2018}) to simulate an 8-joint robot arm. The robot arm has eight segments each of size $\SI{0.0775}{m}$ for a total length of $L=\SI{0.62}{m}$. Each of its eight joints can rotate in $[-\pi/2,\pi/2]\SI{}{rad}$ to position its end-point, a gripper, into any position within the semi-circle spanned by the orientation $[-3\pi/2,2\pi]\SI{}{rad}$. The robot arm is controlled by an 8-dimensional vector which denotes for each joint the desired angle  and it is assumed that this movement is always successful.

During evolution, the simulation environment evaluates the fitness of bottom-level solutions (i.e., the genotypes) as well as the meta-fitness (i.e., the meta-genotypes). A genotype, $\mathbf{g}$, is an 8-dimensional vector of commands, representing the desired angles for each of the joints. The fitness of a genotype $f(\mathbf{g}$), representing the performance of a controller of the robot arm, is the negative variance of the angles made by the joints of the robot arm,
\begin{equation}
 \label{eq: fitness}
 f(\mathbf{g}) = - \frac{1}{8} \sum_{i=1}^{8} (g_i - \bar{g} )^2 \,,
\end{equation}
where $\bar{g} = \frac{1}{8} \sum_{i=1}^{8} g_i$. The fitness discourages zigzag motions, which is energy-efficient, and promotes distributing movement equally among the joints, which is more robust to damages and other perturbations.\footnote{For low-variance motions, the angles will be closer to zero and therefore the maximal absolute change in a joint's angle is reduced.} When the robot arm collides with itself or with the wall, the corresponding genotype is considered unsafe and is not added to MAP-Elites' behaviour-performance map or the database.

Based on a pre-defined damage-set $\mathcal{D}$ that injects damages of different types to the robot arm, the meta-fitness estimates the overall adaptation performance of a meta-genotype $\mathbf{w}$ as follows. First, a batch of genotypes, $\mathcal{B} = \{\mathbf{g}^1,\dots,\mathbf{g}^{N_g}\}$, comprised of 10\% of bottom-level individuals from the behaviour-performance map, is sampled without replacement from the behaviour-performance map. Second, each solution $\mathbf{g} \in \mathcal{B}$ is evaluated on all $d \in \mathcal{D}$, and the final position of the end-point of the robot arm, denoted by $P(\mathbf{g}; d)$, is recorded in each case. Then, the meta-fitness is evaluated as the summed pairwise Euclidian distance of the robots' final positions, averaged across damages in~$\mathcal{D}$:
 \begin{equation}
 \label{eq: meta-fitness}
\mathcal{F}(\mathbf{w}) = \frac{1}{|\mathcal{D}|} \sum_{d \in \mathcal{D}} \sum_{i=1}^{N_g - 1} \sum_{j=i+1}^{N_g} ||  P(\mathbf{g}^i; d) -  P(\mathbf{g}^j; d) || \,.
\end{equation} 
The damage set in meta-evolution included 16 damages, forcing an individual joint $i \in \{1,\dots,8\}$ to get stuck at two different angles, $\theta_{i,1} \sim U(-\pi/2,0)\SI{}{rad}$ and $\theta_{i,2} \sim U(0,\pi/2)\SI{}{rad}$. The damages in  $\mathcal{D}$ therefore cover the legal range of the joint $[-\pi/2,\pi/2]\SI{}{rad}$ across evolution. Partitioning the range into two aims to (i) include both a positive and a negative angle, thereby not biasing the behaviour-performance map to damage of any particular direction; and (ii) reduce the variance of meta-fitness evaluations over time to ensure the meta-fitness is not too deceptive. If a solution collides with itself or the wall, then the corresponding genotype does not contribute to the sum. The highest meta-fitness represents a behaviour-performance map for which all behavioural bins are filled  with valid solutions and for which the solutions are uniformly distributed across the entire semi-circle span of the robot arm regardless of any damage. 
\subsection{Experimental conditions}
\label{sec: experimental-conditions}
We include four baselines, which apply traditional MAP-Elites with a different behaviour space. Contrary to the genotypic space, these are low-dimensional and most avoid unsafe solutions that will be discarded during evolution. \textbf{Position} is a 2-dimensional space denoting the Cartesian coordinate $(x,y)$, normalised from $[-L,L] \times [-L,0]$ to $[0,1]^2$ based on the robot's length $L$, according to $x \gets \frac{x + L}{2L}$ and $y \gets \frac{-y}{L}$. \textbf{Polar} is a 2-dimensional space denoting the polar coordinate $(r,\theta)$, normalised to $[0,1]^2$ based on the length $L$ and the semi-circle range $[\pi,2\pi]\SI{}{rad}$, according to $r \gets r / L$ and $\theta \gets \frac{\theta - \pi}{\pi}$. \textbf{Joint-pair-angle} is the angle in spanned by connecting joint $i-2$ to joint $i$ for all $i \in \{2,4,6,8\}$. Although two consecutive joints can at most both turn $\frac{\pi}{2}\SI{}{rad}$, the Joint-pair-angle ignores the orientation of the previous joint-pair and therefore the angle ranges in $[0,2\pi]\SI{}{rad}$ and is normalised to  $[0,1]$ based on $\theta \gets \frac{\theta}{2\pi}$. \textbf{AngleSum} is a 6-dimensional space that computes the average value for each triplet of bottom-level genes, $\langle 0, 1, 2 \rangle, \langle 1, 2, 3 \rangle, \dots, \langle 6 , 7, 8 \rangle$. The AngleSum can thus be considered as a lower-dimensional formulation of the genotype. \\

We then include a variety of conditions based on QD meta-evolution, which are identified by the \textbf{Meta} prefix and two further suffixes. One suffix denotes the type of feature-map, which can be \textbf{Linear}, \textbf{NonLinear}, or \textbf{Selection} (see Section \ref{subsec: featuremaps}). A second suffix describes the type of parameter control. As a default, if the suffix is not included, then the mutation rate is $0.125$ and there are 5 generations per meta-generation.\footnote{The number of MAP-Elites iterations is divided into distinct generations each of which have a fixed batch-size (see Section \ref{sec: experimental-parameters}).} Four static parameter conditions are included per parameter, where the number of generations per meta-generation is set in $\{5,10,25,50\}$ and the mutation rate is set in $\{0.125,0.25,0.50,1.0\}$, which will be given a distinct suffix (e.g., \textbf{Mutation rate 0.50}, or \textbf{50 generations}). Three dynamic pararameter control settings are included, in which the indicated parameter is allowed to vary within its range, $[1,50]$ for the number of generations and $[0.001,1]$ for the mutation rate. \textbf{Annealing} linearly decreases the parameter $P$ as a function of the number of function evaluations $E$ according to 
$P(E) = m_{P} + (M_P - m_P)\frac{M_{E} - E}{M_E}$, where $m_P$ and $M_P$ are the minimum and maximum of $P$, respectively,  and $M_E$ is the maximal number of function evaluations. \textbf{Endogenous} parameter control adds one gene to the meta-genotype to encode either the number of generations or the mutation rate. Reinforcement learning (\textbf{RL}) follows \cite{Karafotias2014}. It uses the SARSA  algorithm \citep{Rummery1994,Sutton2018b} to select the best action, an interval in the parameter setting, for a given state, data about the progress of evolution. The state of the RL agent is formed from a tree-based discretisation of the continuous observation space, such that states represent different partitions with distinct Q-values \citep{Uther1998}. Our RL setup includes 5 actions, corresponding to 5 equally-spaced bins of the parameter setting. Observations track the progress of meta-evolution, including the maximum, mean, and standard-deviation of meta-fitness, the meta-genotypic diversity, the number of consequent meta-generations the maximal meta-fitness has not improved, and the reward. The reward is the ratio improvement in maximal meta-fitness divided by the added function evaluations.

Finally, analogous to the Meta conditions, we also include Random conditions which perform traditional MAP-Elites but have a randomly initialised feature-map that is not evolved over time. The Random conditions similarly include a suffix for the type of feature-map (Linear, Non-Linear, or Selection).\\
\subsection{Experimental parameters}
\label{sec: experimental-parameters}
We now summarise the parameter settings defining the experiments (see also Table~S1 in Supplemental Materials). Each solution is identified by an 8-dimensional genotype, the angles for each of the joints of the robot arm. The mutation operator is a random increment or decrement to the gene, with a step of $0.025$, with a mutation rate of $0.125$ such that each child solution is on average differing in one gene from its parent. The maximal map coverage allowed for each algorithm is 4,096 solutions. In the grid-based geometry of MAP-Elites, assuming equal number of bins per dimension, this allows 64 bins for 2D spaces (Position and Polar conditions), 8 bins for 4D spaces (Meta conditions), and 4 bins for 6D spaces (AngleSum condition). All conditions have 400 bottom-level individuals per generation and an initial population of 2,000 randomly generated individuals. The large budget of 100,000,000 function evaluations allows evolution to run its full course.

For Meta conditions, the meta-population size is set to $\lambda=5$ based on the trade-off between global search and convergence. Due to the experimental conditions (see Section \ref{sec: experimental-conditions}) there are $N_b=14$ base-features and $D=4$ target-features, so the meta-genotype comprises a total of 56 genes, except for the non-linear feature-maps, which require 182 ($N_b \times N_h  + N_h \times D + 2$) genes based on $N_h=10$ hidden units. For linear feature-maps, a normalisation range of $[0.20,0.80]$ is set based on the empirical range of the target-features. For non-linear feature-maps, the scaling factor in Eq. \ref{eq: nonlinfm} is set to $\alpha_s = 30$  which yields a theoretical range of $[-30,30]$ and an empirical range of $[-5,5]$. This setting allows covering the entire output range $[0,1]^D$ with sufficient precision and it is the smallest setting of $\alpha_s$ for which extreme feature values (close to 0 or 1) are equally frequent as the middle feature values. Endogenous parameter control adds one gene to the meta-genotype. Initialisation of CMA-ES starts with the middle of the parameter range as the mean and a third of the range as the standard-deviation. The meta-genotype parameters' ranges are $[-1,1]$ for non-linear feature-maps and $[0,1]$ otherwise. Further parameters of CMA-ES use the default settings. The database has a capacity of just below 5 million, allowing 3 bins for all 14 features for a bin width of $\delta=1/3$, and initially, the number of solutions per coarse bin is set to $k=5000$.
\begin{figure*}[htbp!]
\centering
\subfigure[Map coverage]{\includegraphics[height=0.25\linewidth]{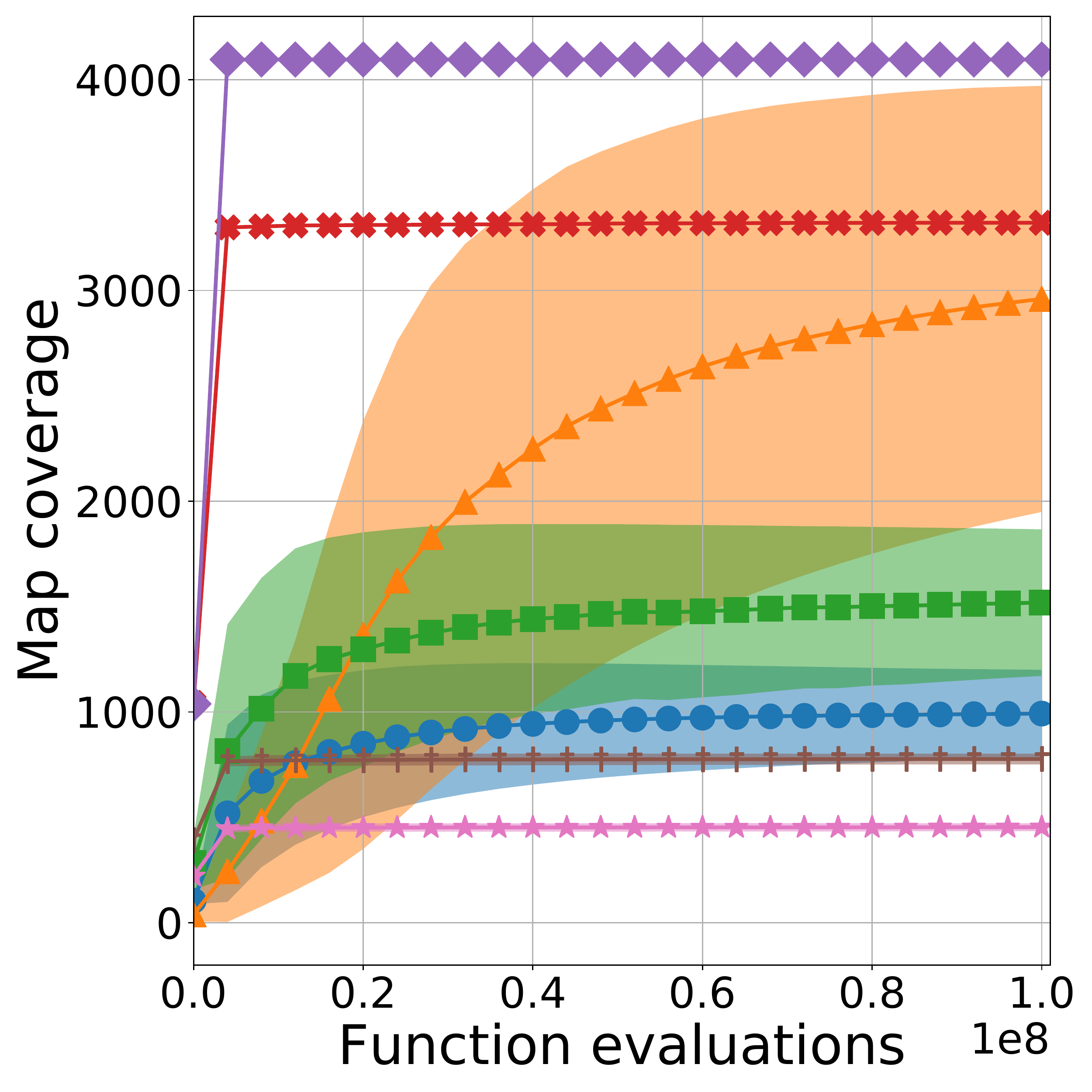}}               \label{fig: coverage}
\subfigure[Global fitness]{\includegraphics[height=0.25\linewidth]{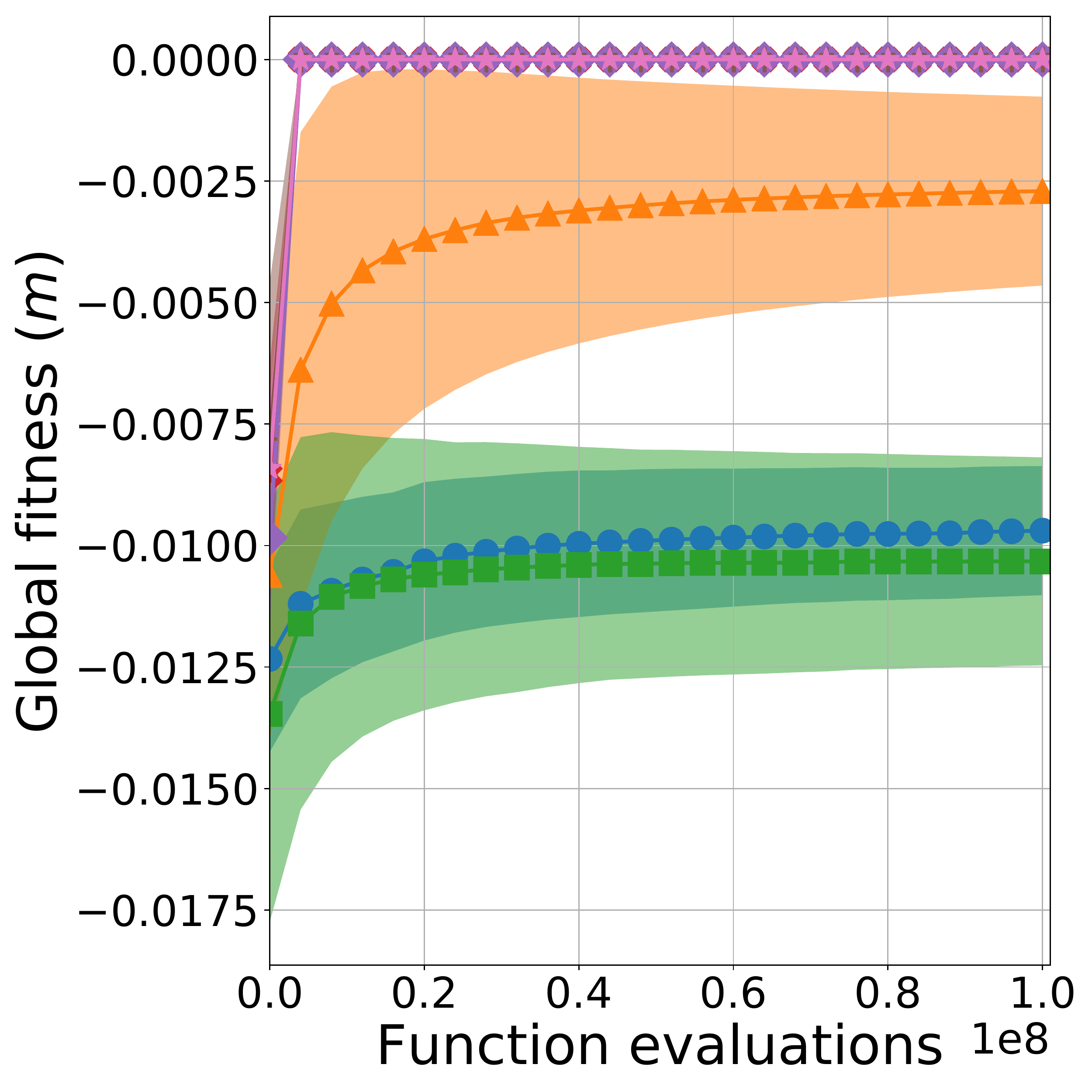}}   \label{fig: globalperformance}
\subfigure[Average fitness]{\includegraphics[height=0.25\linewidth]{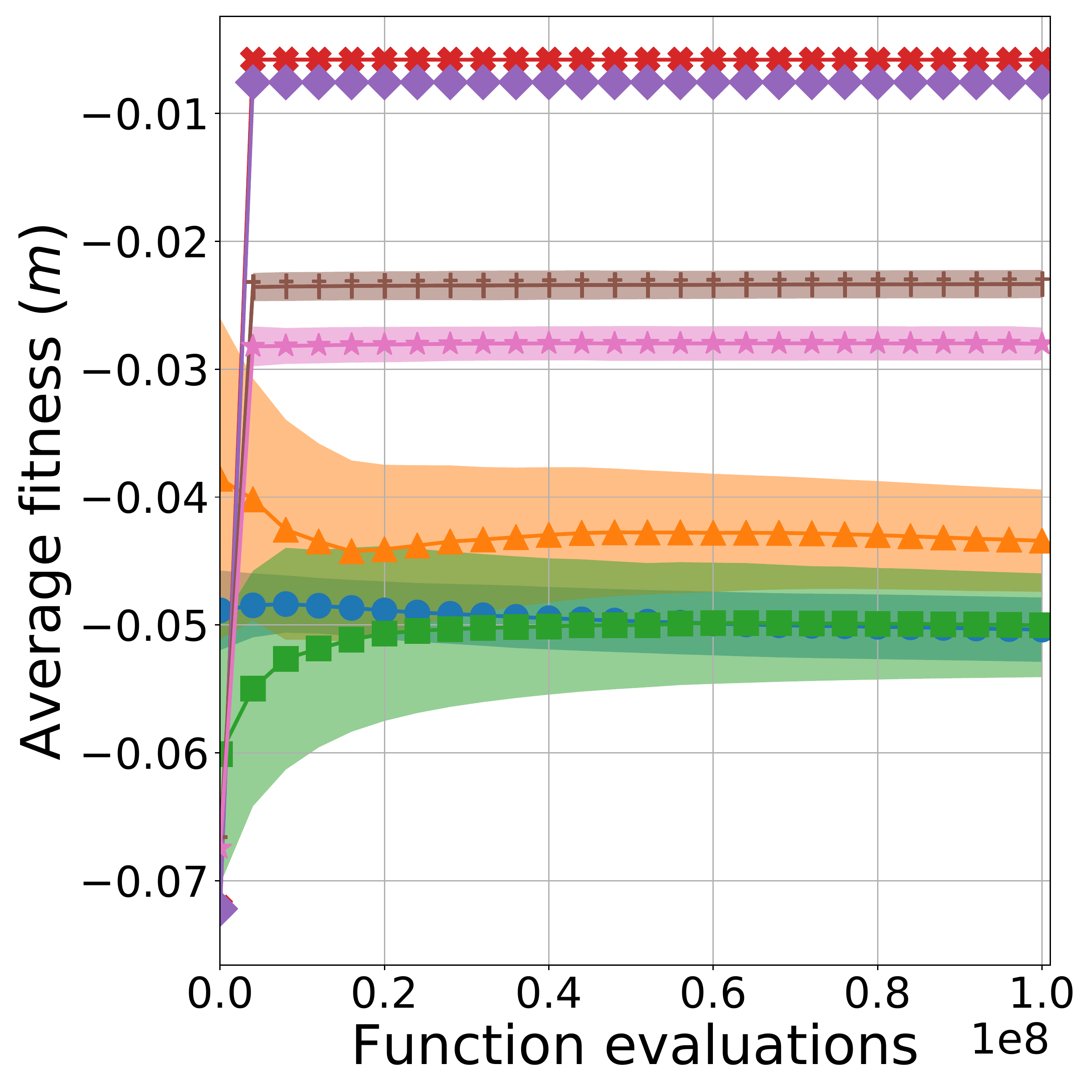}}   \label{fig: averageperformance}
\includegraphics[width=0.35\linewidth]{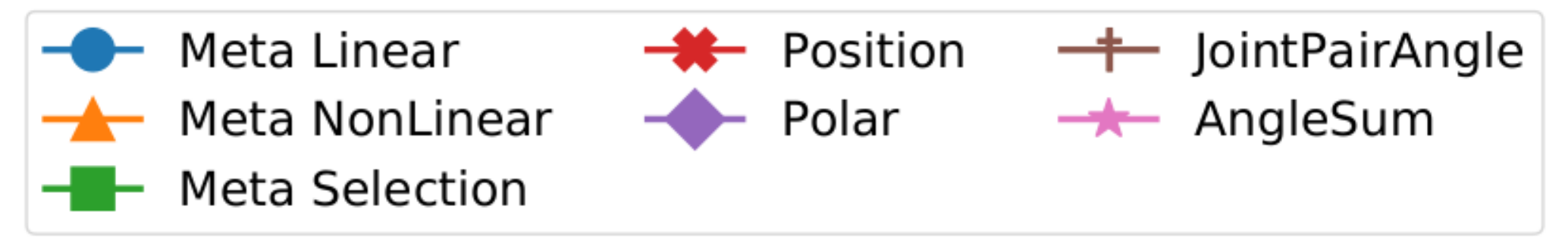}
\caption{Evolution of the following map quality metrics (Mean $\pm$ SD, aggregated across replicates) over the function evaluations: (a) map coverage, the number of solutions in the behaviour-performance map; (b) global fitness, the highest bottom-level fitness in the map; (b) average fitness, the mean bottom-level fitness in the map. For Meta-conditions, Mean and SD are aggregated over replicates and the behaviour-performance maps, and default hyperparameters are used (mutation rate 0.125 and 5 generations per meta-generation). Fitness is the negative variance of the angles of the robot arm when normalised to $[0,1]$ such that $0$ represents $\SI{-\pi/2}{rad}$ and $1$ represents $\SI{-\pi/2}{rad}$. } \label{fig: evolution}
\end{figure*}
\section{Results}
We now demonstrate the improved meta-evolution system on a planar 8-joint robot arm. First, we evaluate the quality of the evolved maps by their coverage of the behaviour space and their fitness. We then investigate the evolution of meta-fitness depending on the type of feature-map and parameter control. Finally, we evaluate the experimental conditions on a damage recovery test. Further meta-fitness and damage recovery comparisons can be found in Supplemental Materials.
\subsection{Quality of evolved maps}
Among the Meta-conditions, the non-linear feature-map obtains the highest coverage of 3000 solutions, compared to 1000 and 1500 for linear and feature-selection feature-maps, respectively (see Figure~\ref{fig: evolution}a) -- this despite random non-linear feature-maps having extremely low coverage, often of only 1 or 2 solutions, as they output a similar value for a large proportion of its inputs. This is explained by the success in optimising the meta-fitness in Eq.~\ref{eq: meta-fitness}, which is strongly related to the coverage. The highest coverages are obtained by Polar and Position (all 4096 and around 3300) and the lowest by JointPairAngle and AngleSum (around 800 and 500).

All conditions are able to find at least one controller with near maximal fitness ($-0.01$ is the lowest maximum observed; see Figure~\ref{fig: evolution}b). The average fitness of Meta-conditions is relatively low, ranging from $-0.05$ to $-0.04$, whereas Polar and Position range from $-0.01$ to $-0.005$ and AngleSum and JointPairAngle range from $-0.03$ to $-0.02$ (see Figure~\ref{fig: evolution}c). This is likely due to a combination of two factors: (i) the features may depend strongly on the angle rather than end-position and so certain behavioural bins can only have low fitness; (ii) compared to other angle-based behaviour-spaces (AngleSum and JointPairAngle), the coverages are higher and still increasing, which slows down fitness improvements.\\
\begin{figure}[htbp!]
\centering
\includegraphics[width=0.65\linewidth]{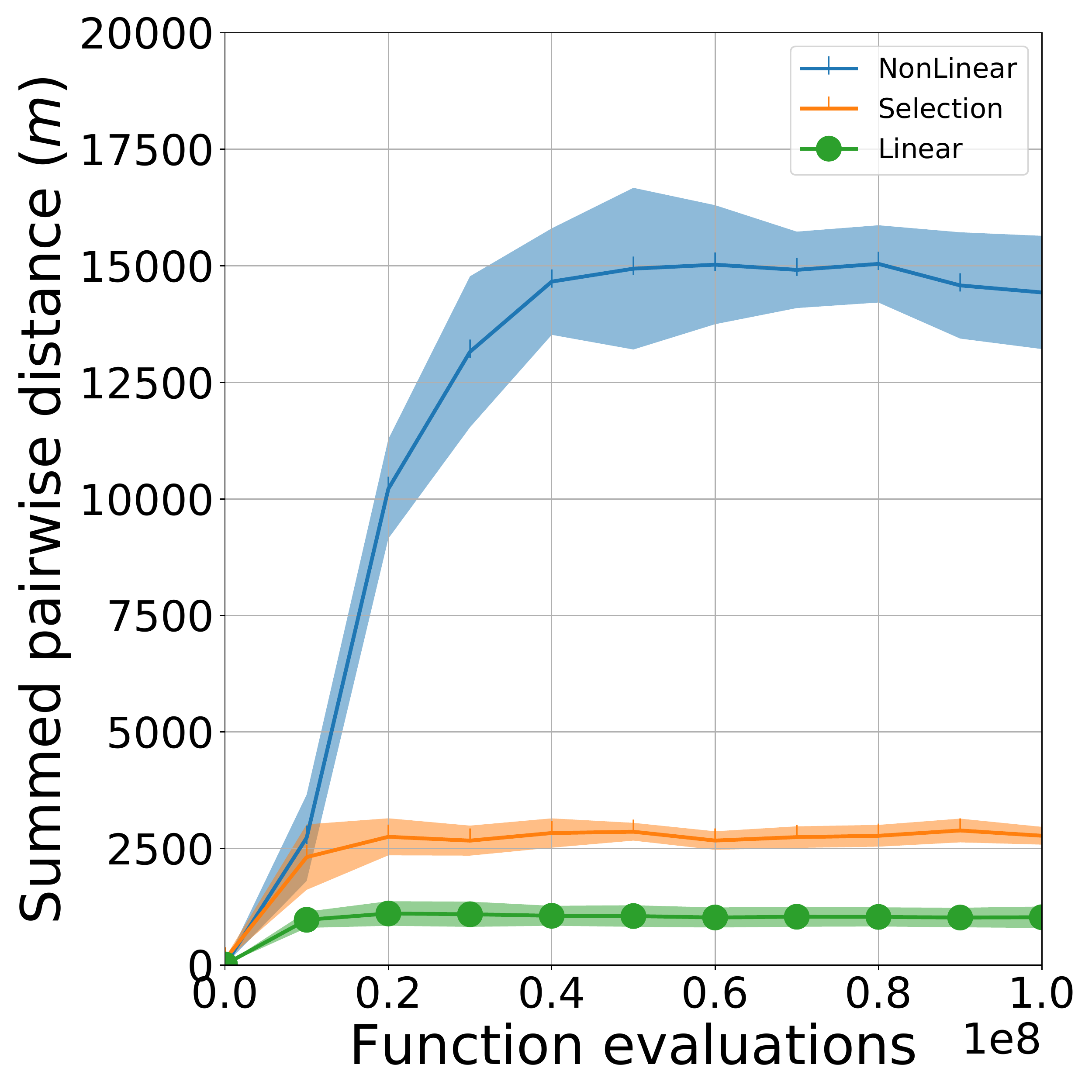}  \label{fig:  feature-maps_epochs}
\caption{Effect of the feature-map on meta-evolution. $x$-axis represents the number of function evaluations and $y$-axis represents the meta-fitness, the summed pairwise distance across 10\% of solutions in the map averaged across the damages in which it is assessed. The average meta-fitness in the $\lambda=5$ behaviour-performance maps in the meta-population is first computed and then this average is aggregated over 5 replicates as Mean $\pm$ SD. Default hyperparameters are used (mutation rate 0.125 and 5 generations per meta-generation). } \label{fig: meta-fitness}
\end{figure}
\subsection{Effect of feature-maps on meta-fitness} 
\label{subsec: feature-maps}
The optimisation of meta-fitness, the summed pairwise distance among safe solutions, is strongly dependent on the feature-map (see Figure \ref{fig: meta-fitness}). While linear and feature-selection feature-maps enable rapid improvements in meta-fitness early on, they then stagnate on a plateau for the rest of meta-evolution. By contrast, non-linear feature-maps start slowly but between 10 million and 20 million function evaluations they improve rapidly. Non-linear feature-maps reach the highest meta-fitness of around $\SI{15000}{m}$ in summed pairwise distance, a 6-fold improvement over feature-selection with meta-fitness of around $\SI{2700}{m}$  and a 15-fold improvement over the linear feature-map with  meta-fitness of around $\SI{1000}{m}$. This illustrates the trade-off of high-complexity functions: they can in principle represent the required function to optimise an objective but they require more data.
\subsection{Effect of parameter control on meta-fitness} 
\label{subsec: parameter control}
The best parameter control method varies depending on the feature-map and the type of parameter but RL is consistently top-ranked. For linear feature-maps, RL obtains the highest scores by a large margin, with a final meta-fitness of $1088.7 \pm 373.4$  when controlling the number of generations per meta-generation and $1640.5 \pm 340.1$ when controlling the mutation rate. In these cases, RL converges to around 25 generations and around 0.75 mutation rate in all of the replicates. For non-linear feature-maps, static control with a lower number of generations of 5--10 and higher mutation rates 0.25--1 are preferable, yielding meta-fitness of $14600$--$14700$ and $16300$--$16500$, respectively. For feature-selection feature-maps, 50 generations and mutation rate 0.50 yield the highest mean meta-fitness of $3817.6 \pm 520.2$  and $5067.7 \pm 263.2$, respectively. These settings of the number of generations per meta-generation can be intuitively understood, because a smaller number of generations per meta-generation is preferred when the space of feature-maps is larger (non-linear feature-maps) and a higher number of generations per meta-generation is preferred otherwise (feature-selection feature-maps) as its provides more reliable estimates of a particular meta-genotype's meta-fitness. In sum, control of the mutation rate can yield up to a $90\%$ improvement in the meta-fitness while control of the number of generations per meta-generation yields up to $40\%$ improvement. 
\subsection{Recovery from a priori unknown damages}
We now assess the robot arm on the same simulation environment albeit with a different damage set than in meta-evolution. Rather than putting the joint stuck at a particular angle, the damage set $\mathcal{D}_{\text{test}}$ applies to each angle of the joint a particular offset $\epsilon \in [-\pi,\pi]\SI{}{rad}$, such that the resulting angle of each joint is $\theta_i \gets \max(-\pi/2, \min(\pi/2, \theta_i + \epsilon))\SI{}{rad}$. The included offset angles are the 20 equally spaced segments in the range excluding zero, i.e. $\{-1, -0.9,\dots, -0.1, 0.1, \dots, 0.9, 1 \} \SI{\pi}{rad}$. Given the observed evolutionary trends, we first investigate QD meta-evolution with non-linear feature-map.

\begin{figure}[htbp!]
\centering
\subfigure[Joint 1]{\includegraphics[width=0.30\linewidth]{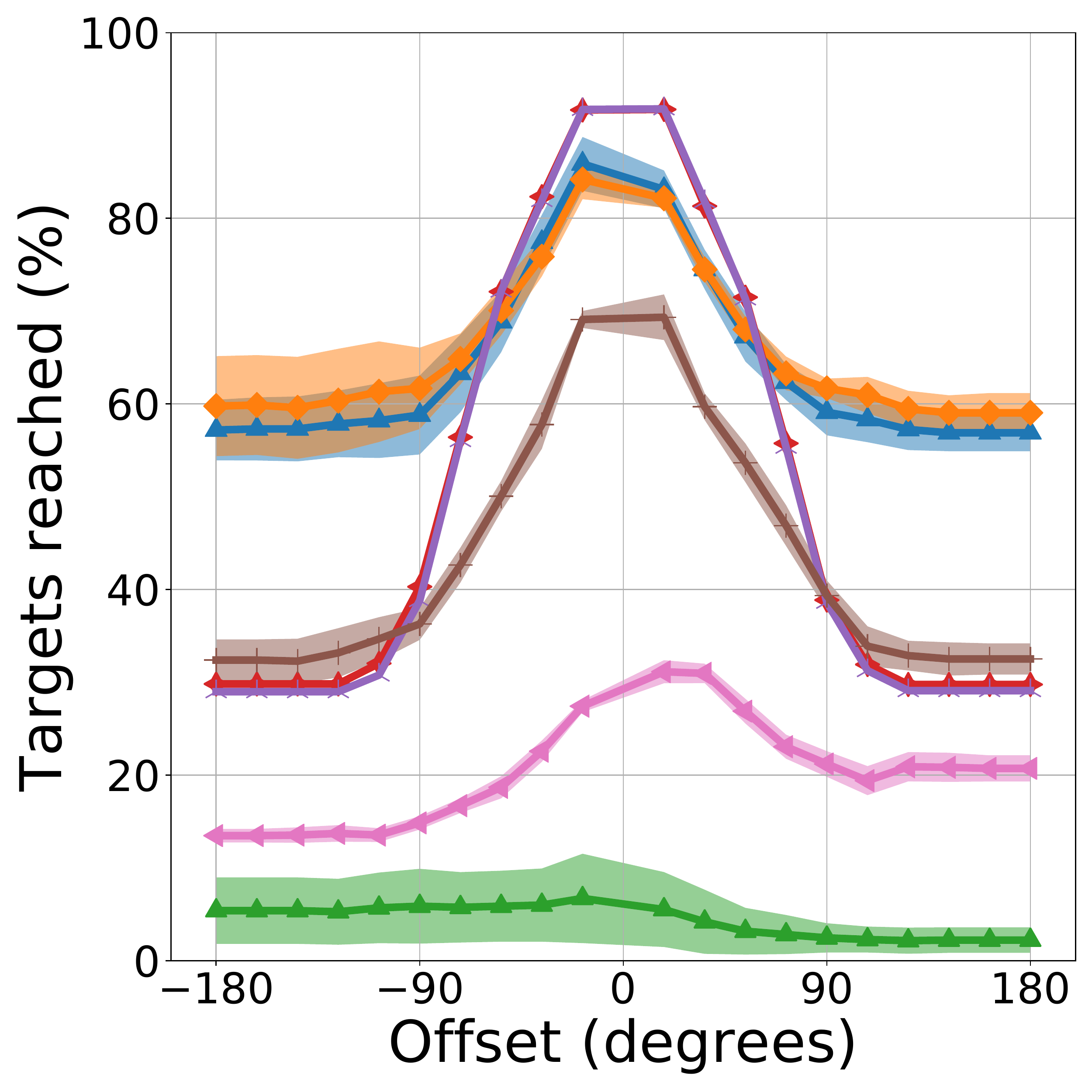}}  
\subfigure[Joint 8]{\includegraphics[width=0.30\linewidth]{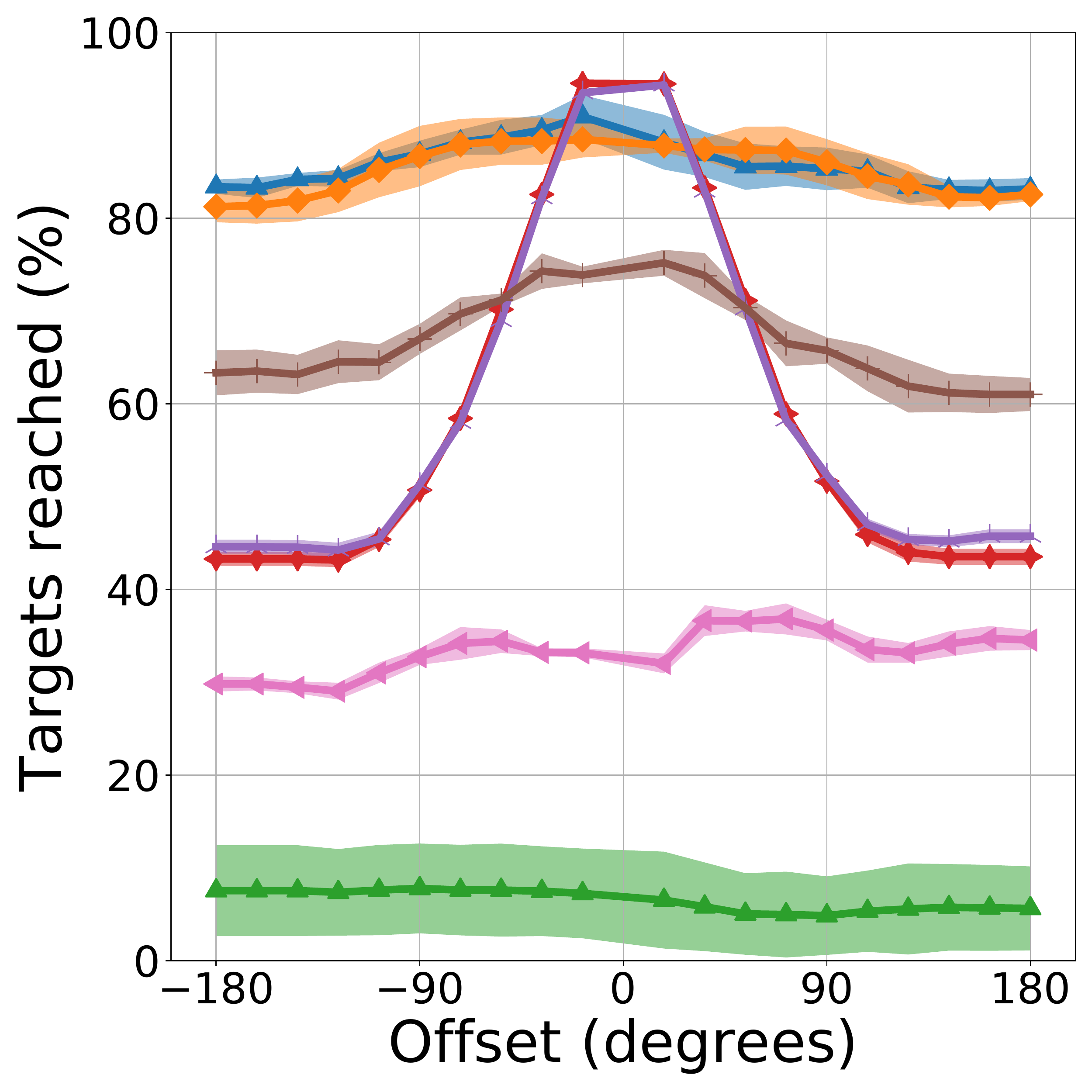}}  
\subfigure[Joint 8]{\includegraphics[width=0.30\linewidth]{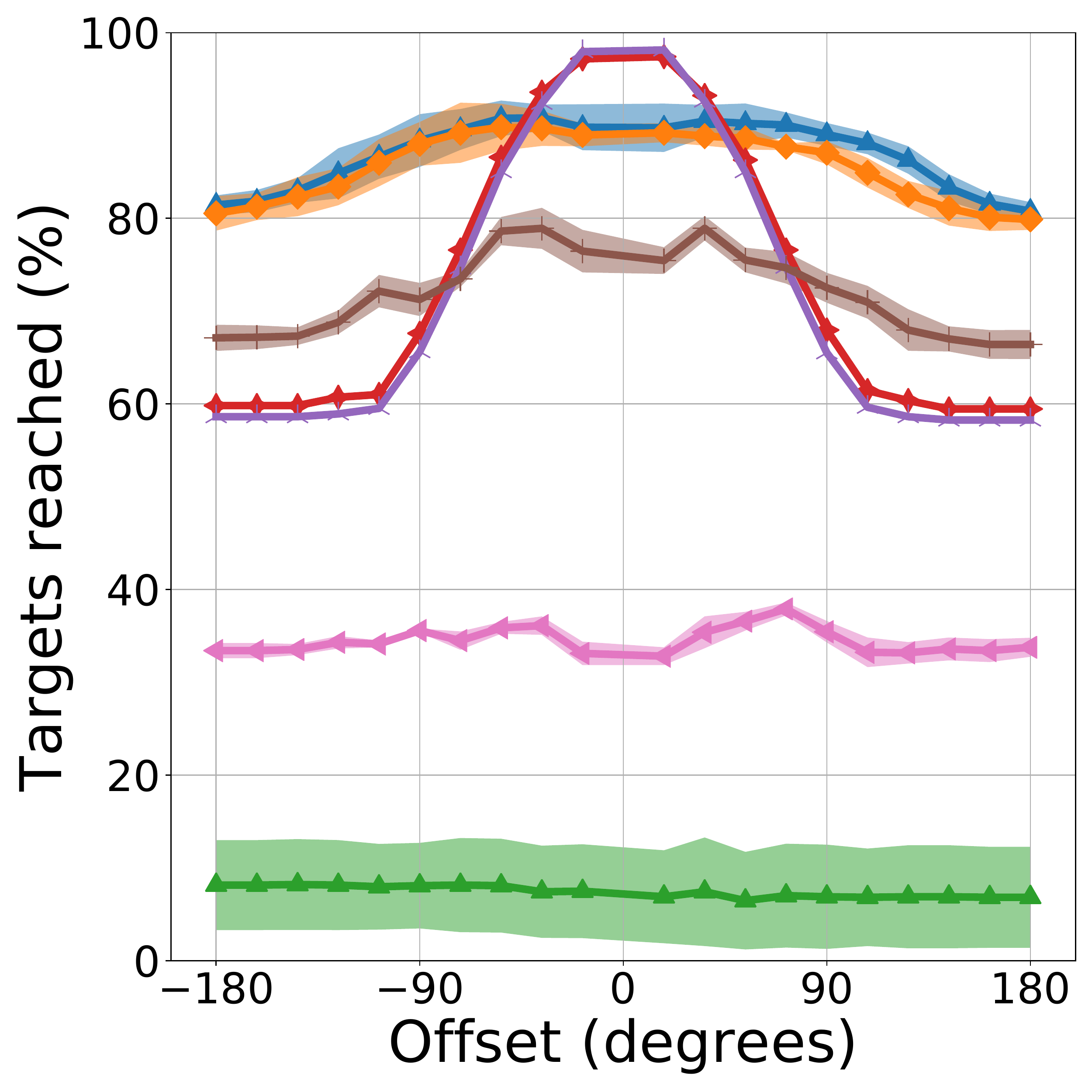}} 
\includegraphics[width=0.90\linewidth]{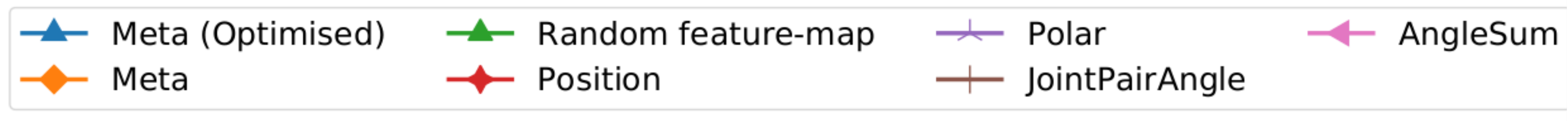}
\caption{Test on unseen damages that offset the joint by a particular angle. The $x$-axis represents the offset in $[-180,180]$ degrees and the $y$-axis represents the percentage of targets reached within the semi-circle span of the robot. For each offset the Mean $\pm$ SD is  aggregated over 5 replicates. For Meta, the behaviour-performance map is formed from the mean meta-genotype (see $\mathbf{m}$ in Eq.~\ref{eq: multivariate normal}) and the default hyperparameters are used (mutation rate 0.125 and 5 generations per meta-generation). \textbf{Optimised} indicates the best setting from parameter control (i.e., mutation rate 0.25). }\label{fig: testperformance}
\end{figure}

For high-severity damages such as offsets of 180 degrees on the upper joint, QD meta-evolution reaches at least 60\% of the target positions in its semi-circle span (see Figure \ref{fig: testperformance}), whereas Polar and Position, the highest performers of traditional MAP-Elites algorithms, have a minimum of 30\%. For smaller offsets below 45 degrees, Polar and Position behaviour-performance maps achieve the highest reach of up to 95\% of targets followed closely by QD meta-evolution which has a reach of 80--90\%. MAP-Elites with a random non-linear feature-map fails completely with only 1\% of targets being reached. \\

To assess statistical significance, we aggregate data of all joint damages and replicates and assess the mean and standard-deviation, statistical significance, and the effect size (see Table \ref{tab: significance}). QD meta-evolution outperforms all other included algorithms with statistical significance, $p<0.001$, and large effect size, Cliff's delta greater than 0.5. \\

\begin{table}[htbp!]
\centering
\caption{Summary statistics of test on unseen damages. For each condition, For each meta-condition, we show the percentage of targets reached (Mean $\pm$ SD) within the semi-circle span of the robot arm and, to assess the effect compared to QD meta-evolution, the Wilcoxon rank-sum test's significance value and Cliff's delta as an effect size. Bold highlights large effect sizes. For Meta, the behaviour-performance map is generated from the mean meta-genotype (see $\mathbf{m}$ in Eq. \ref{eq: multivariate normal}) and default hyperparameters are used (mutation rate 0.125 and 5 generations per meta-generation). \textbf{Optimised} indicates the best setting from parameter control (i.e., mutation rate 0.25).} \label{tab: significance}
\begin{tabular}{l l l l}
\toprule
\textbf{Condition} & Targets reached ($\%$) & Significance & Cliff's delta  \\ \hline
Meta (Optimised) & $79.16 \pm 11.38$ & \quad /  & \quad  / \\ 
Meta & $79.10 \pm 10.66$ & $p=0.338$ & $0.03$ \\
Random NonLinear & $5.94 \pm 4.64$ & $p<0.001$ & $\mathbf{1.00}$ \\ 
Position & $58.81 \pm 19.39$ & $p<0.001$ & $\mathbf{0.57}$\\
Polar & $57.87 \pm 19.49$  & $p<0.001$ & $\mathbf{0.60}$ \\ 
JointPairAngle & $60.53 \pm 13.38$ & $p<0.001$ & $\mathbf{0.71}$\\ 
AngleSum & $30.04 \pm 6.63$  & $p<0.001$ & $\mathbf{1.00}$\\ 
\bottomrule
\end{tabular}
\end{table}

A similar comparison of Meta-conditions demonstrates that Meta NonLinear significantly outperforms other Meta-conditions. Meta NonLinear and Meta Linear generalise as expected but Meta Selection has selected features that are optimised for the training damage set but generalise poorly to the test damage set. After test damages, only  5\% to 25\% of solutions in the archive remains safe for Meta Selection, whereas Meta Linear and Meta NonLinear retain 25\% to 100\% safe solutions (roughly 50--250, 250--1000 and 1000--3500 solutions, respectively). The high test-performance of Meta NonLinear is not due to overlap in train- and test-damages, as we observe comparable test performances after dropping the damage set from meta-evolution such that only behavioural diversity and not environmental diversity is included in meta-fitness evaluations. The near equivalence in test-performance with and without damages may be explained by the strong relation between behavioural diversity and environmental diversity (see e.g., \cite{Bossens2020}).
\section{Conclusion}
In quality-diversity algorithms such as MAP-Elites, a key challenge is to design a behaviour space and a set of evolutionary operators. We present a novel QD meta-evolution system that automatically determines the behaviour space and evolutionary parameters for the MAP-Elites algorithm. The system extends Meta-evolution with CMA-ES \citep{Bossens2020a}, a prior quality-diversity meta-evolution algorithm, with three modifications: (i) the database is reformulated to selectively maintain solutions based on their contribution to quality and diversity; (ii) feature-maps that can represent arbitrary functions of the base-features and the meta-genotype; and (iii) the use of parameter control to adjust the mutation rate and the number of generations per meta-generation for the MAP-Elites algorithms.

Improved QD meta-evolution is demonstrated on an 8-joint planar robot arm studying various feature-maps (linear, non-linear, and feature-selection) as well as various parameter control strategies (static, endogenous, reinforcement learning, and annealing). In evolution, non-linear feature-maps yield a 15-fold improvement over the traditional linear transformation used in prior meta-evolution work although they require more function evaluations. Such feature-maps cannot be hand-crafted and are unique representations customised to a meta-level objective. Feature-selection provides a 3-fold improvement over linear feature-maps without requiring more function evaluations. Parameter control yields up to 90\% meta-fitness improvement for the mutation rate and up to 40\% meta-fitness improvement for the number of generations per meta-generation. Reinforcement learning consistently ranks among top parameter control methods and therefore is recommended as a generic method to avoid tuning. Subsequent damage-recovery tests demonstrate that behaviour-performance maps evolved by QD meta-evolution with non-linear feature-map can reach diverse target positions, regardless of the severity of the damage sustained by the robot arm.


\section*{Acknowledgements}
This work was supported by EPSRC under the New Investigator Award grant, EP/R030073/1 (Tarapore). The authors  acknowledge the IRIDIS High Performance Computing Facility and thank Arnold Benedict for initial work on the simulator.

\bibliographystyle{ACM-Reference-Format}
\bibliography{sample-bibliography} 

\end{document}


\title{Supplemental Materials for \\ \textit{On the use of feature-maps and parameter control for improved quality-diversity meta-evolution}}

\author{David M. Bossens\\
	University of Southampton\\
	\texttt{D.M.Bossens@soton.ac.uk} \\
	\And
	Danesh Tarapore \\
	University of Southampton\\
}

\maketitle
\thispagestyle{empty}

\section{Experimental parameters}
For convenience, Table~\ref{tab: evolutionparameters} includes all the parameter settings for the experimental setup.

\begin{table}[htbp!]
\centering
\caption{Parameter settings for evolution. Top half shows settings common to all conditions while bottom half shows settings for Meta-conditions.}  \label{tab: evolutionparameters}
\begin{tabular}{l  p{4.4cm}}
\toprule
\textbf{Parameter}  & \textbf{Setting} \\ \hline
Genotype  ($\mathbf{g}$) & discretised in $[0,1]^{8}$  \\
Mutation rate   &   $0.125$ (unless otherwise \newline indicated) \\
Mutation type &   random increment/decrement with step of $0.025$ \\
Maximal map coverage & 4,096 solutions \\
Function evaluations & 100,000,000 \\ 
Batch size per generation &   400 bottom-level individuals \\
Initial population ($p$) &  2,000 bottom-level individuals \\
\hline
Meta-population size ($\lambda$) & 5 \\
Meta-genotype    ($\mathbf{w}$) & $[-1,1]^{182}$ for non-linear feature-map \newline $[0,1]^{56}$ otherwise \\
Number of base-features ($N_b$) & 14\\
Number of target-features ($D$) & 4\\
Normalisation range ($[m,M]$) & $[0.20,0.80]$ (linear feature-maps) \\
Number of hidden units ($N_h$) & 10 (non-linear feature-maps)\\
Sigmoid scaling factor ($\alpha_s$)  & 30 (non-linear feature-maps)\\
Database settings  & initial $k=5000$; bin width $\delta=1/3$; capacity $3^{14}$ (just below 5 million) \\
\bottomrule
\end{tabular}
\end{table}
\newpage
\section{Meta-fitness development}
\begin{figure*}[htbp!]
\centering
\subfigure[Linear]{\includegraphics[width=0.31\linewidth]{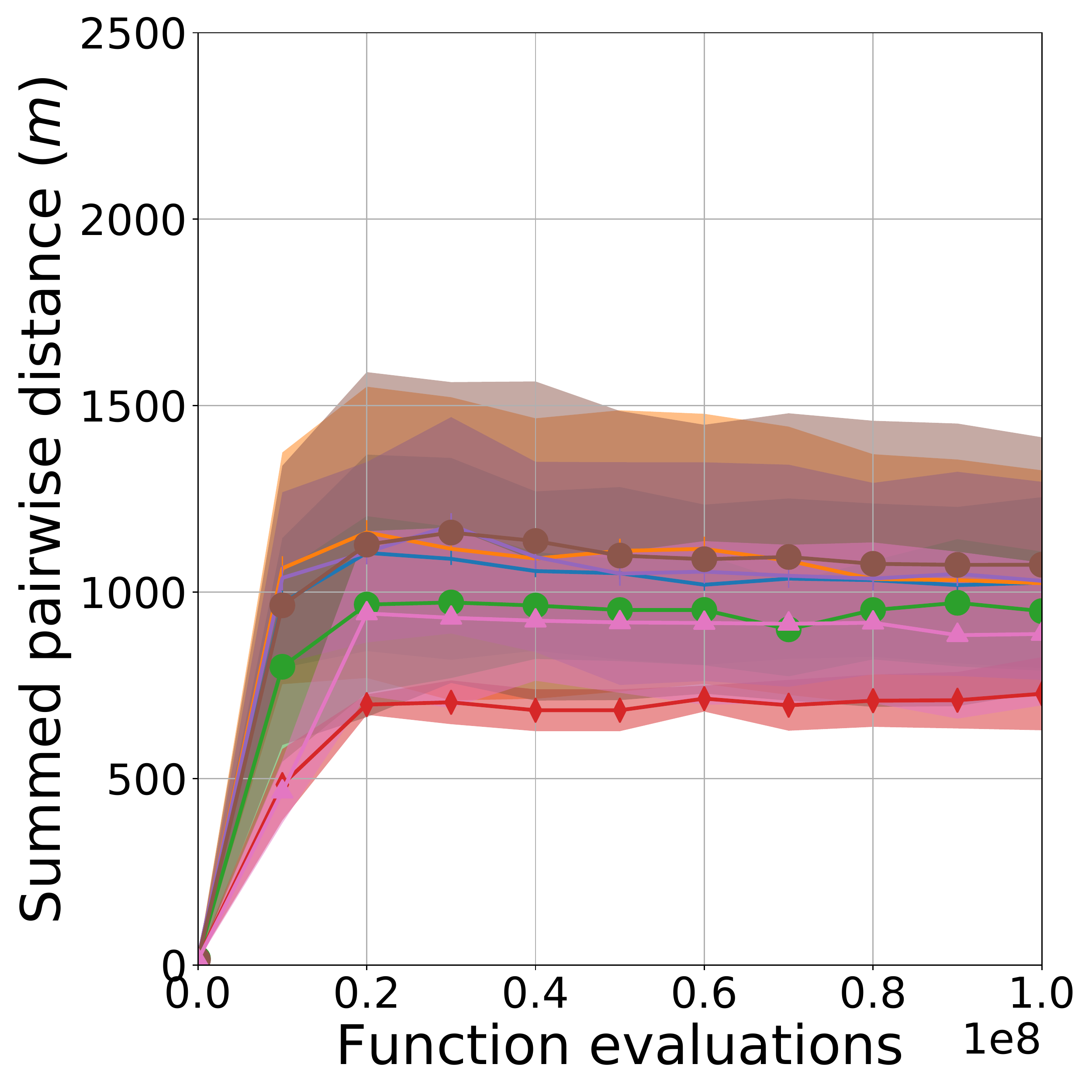}}   \label{fig: linfm_epochs}
\subfigure[Non-linear]{\includegraphics[width=0.31\linewidth]{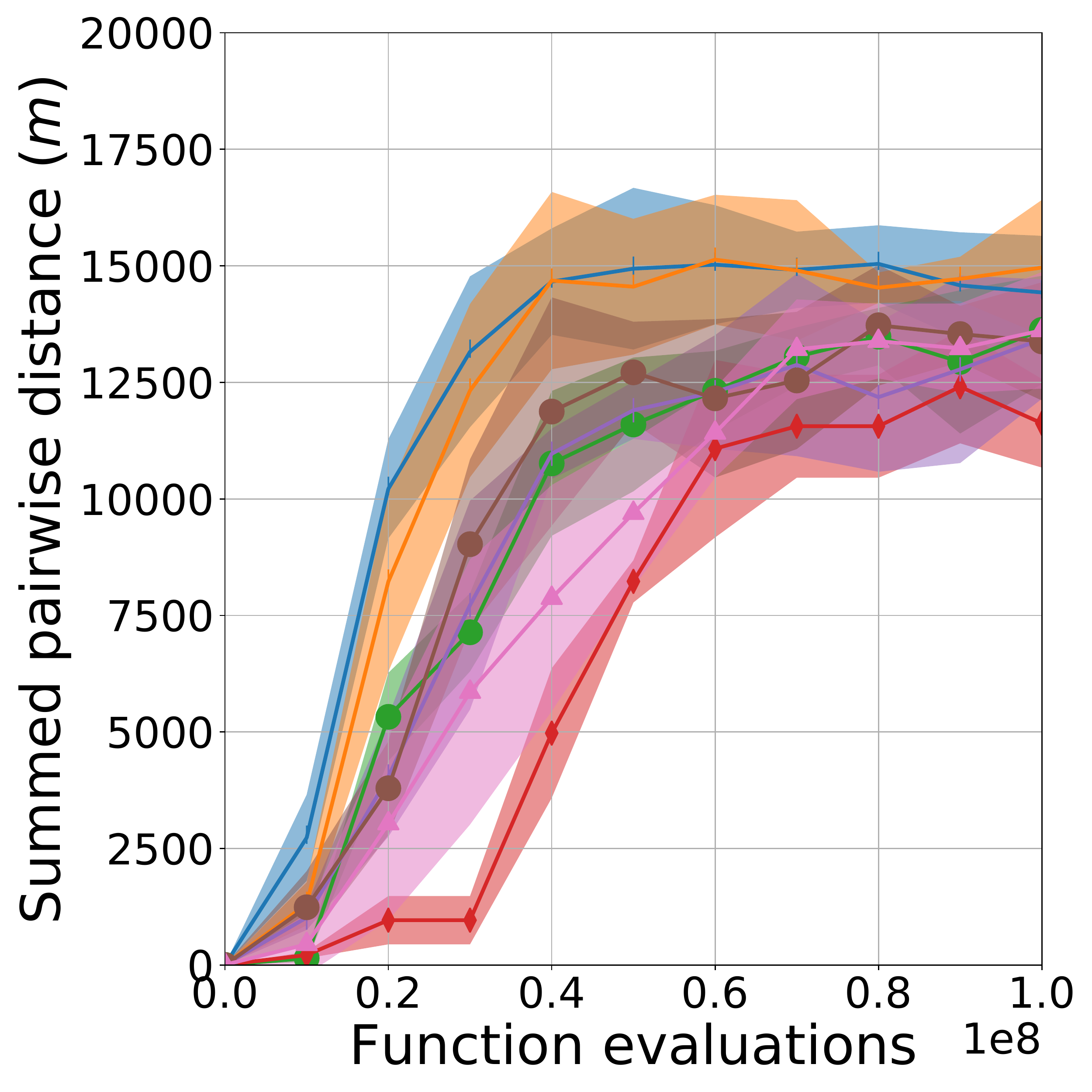}}   \label{fig: nonlinfm_epochs}
\subfigure[Feature selection]{\includegraphics[width=0.31\linewidth]{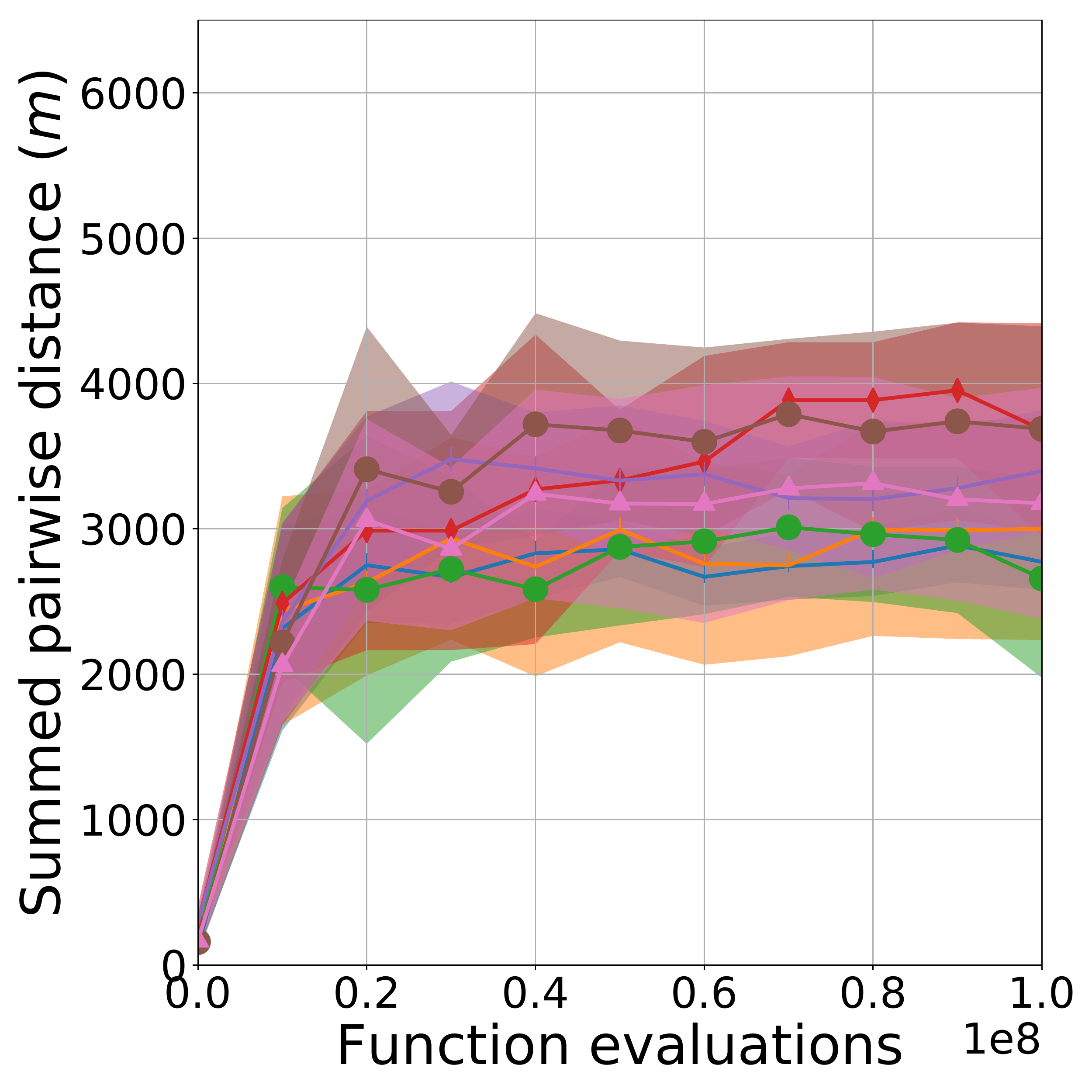}}               \label{fig: selection_epochs}
\includegraphics[width=0.90\linewidth]{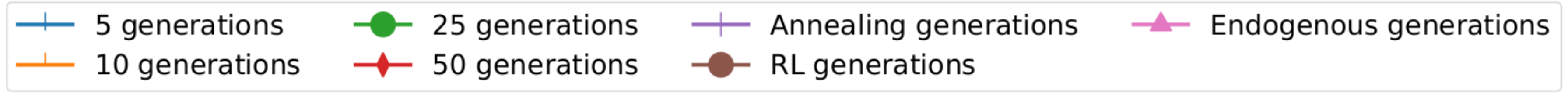}
\caption{Analysis of parameter control of the bottom-level epochs, the number of generations per meta-generation. The $x$-axis represents the number of function evaluations and the $y$-axis represents the meta-fitness, the summed pairwise distance across 10\% of solutions in the map. To compute a single number and indicate variability, Mean $\pm$ SD of the population average meta-fitness is aggregated over 5 replicates.} \label{fig: epochs-control}
\end{figure*}

\begin{figure*}[htbp!]
\centering
\subfigure[Linear]{\includegraphics[width=0.31\linewidth]{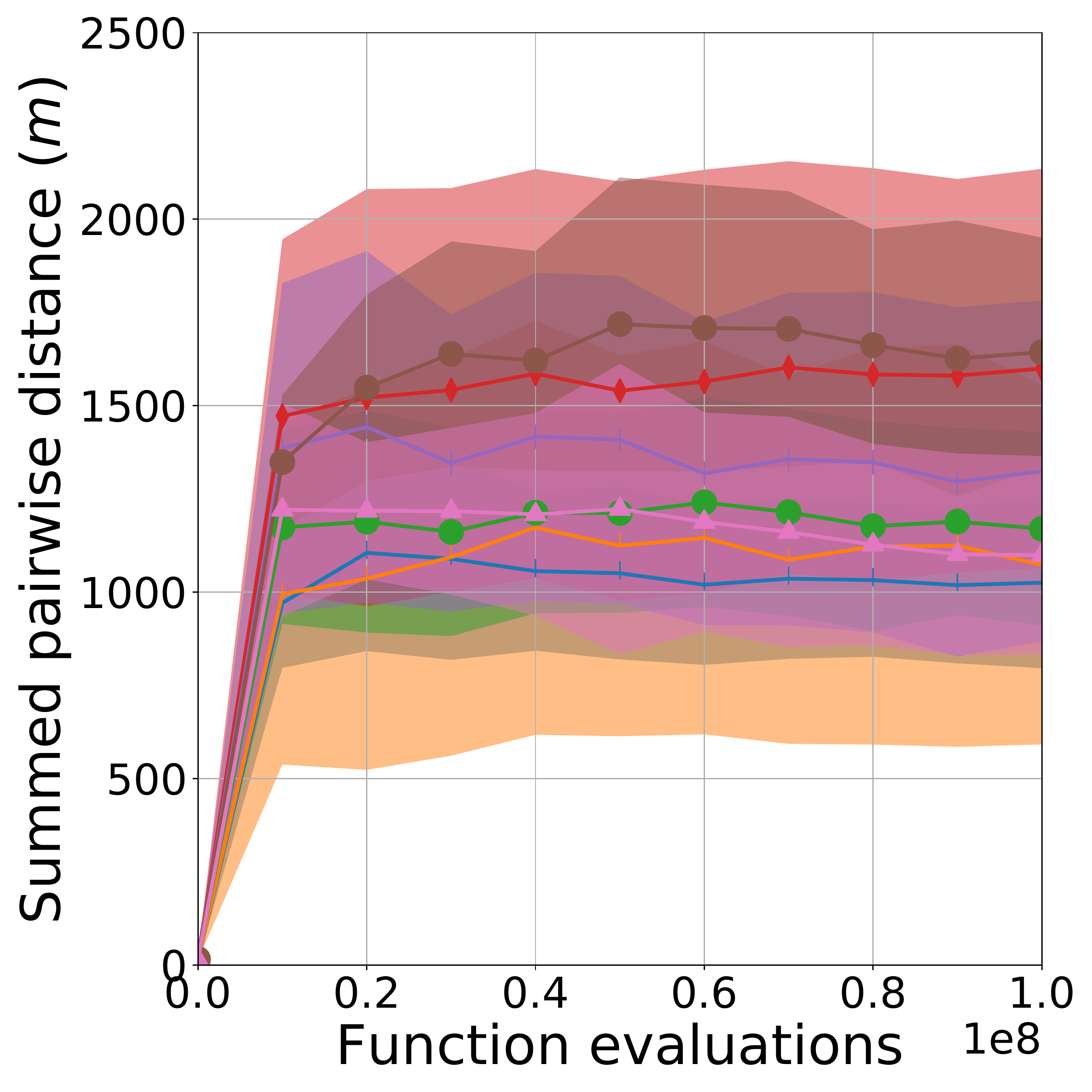}}   \label{fig: linfm_mutationrate}
\subfigure[Non-linear]{\includegraphics[width=0.31\linewidth]{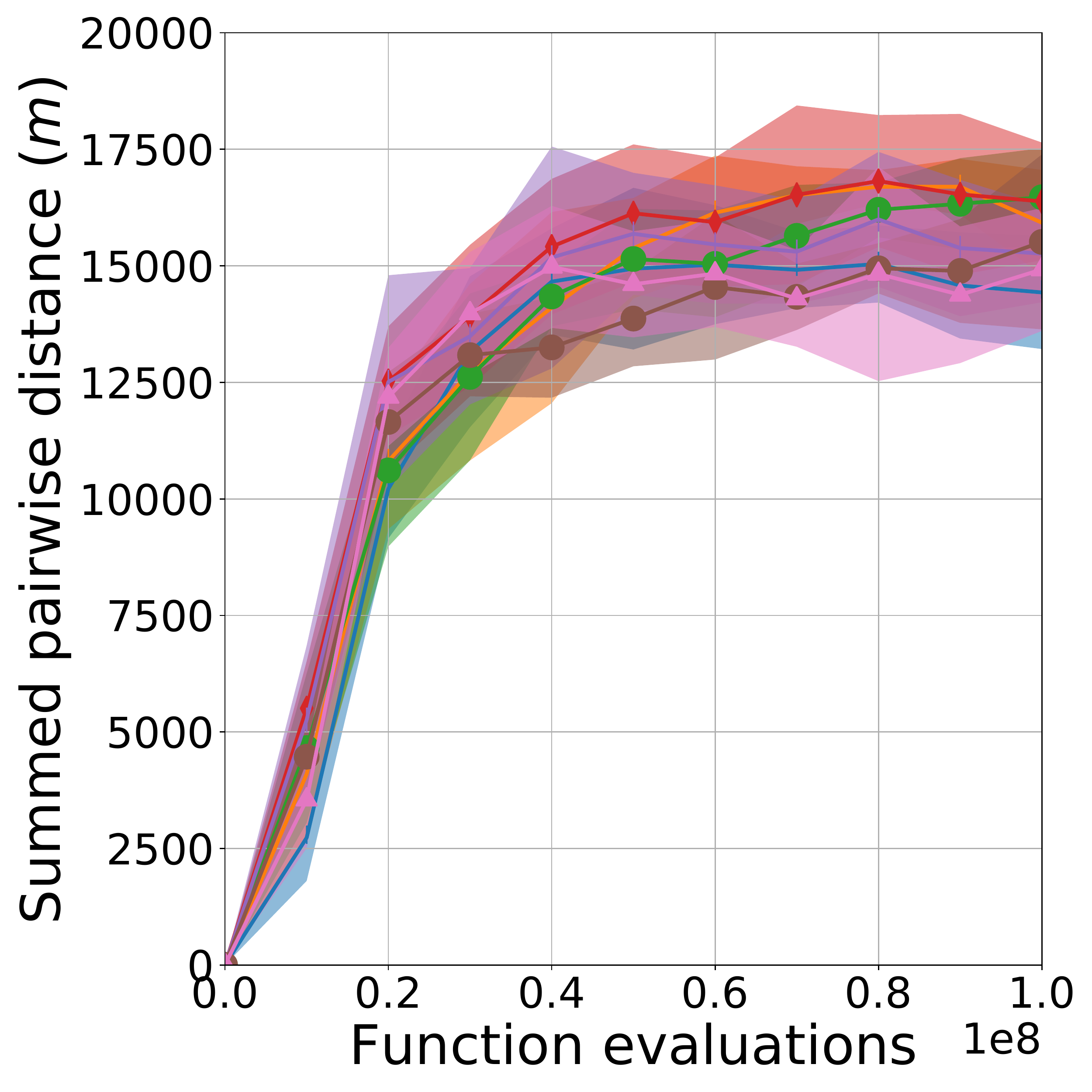}}   \label{fig: nonlinfm_mutationrate}
\subfigure[Feature selection]{\includegraphics[width=0.31\linewidth]{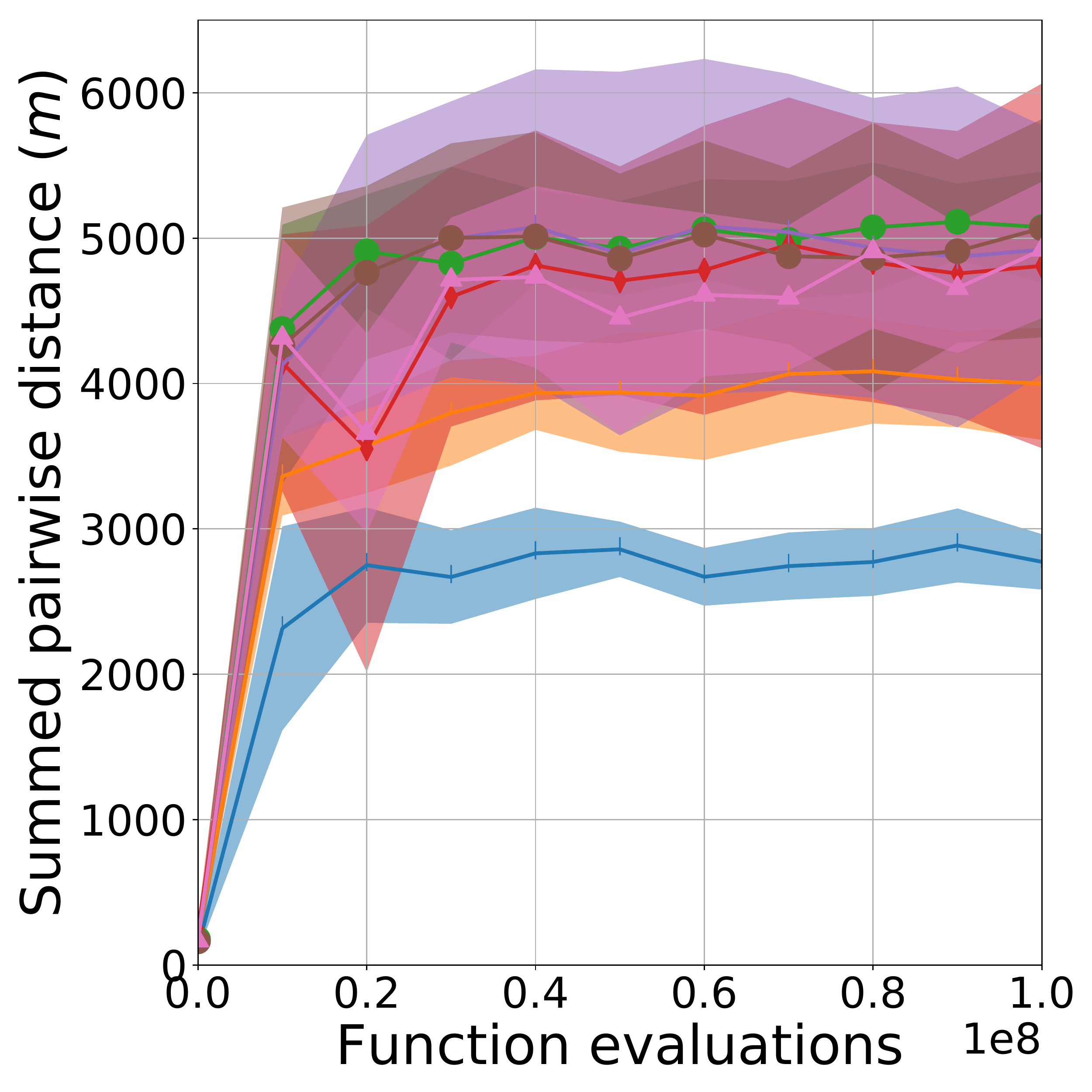}}               \label{fig: selection_mutationrate}
\includegraphics[width=0.90\linewidth]{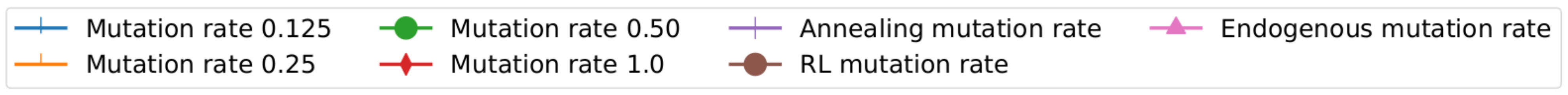}
\caption{Analysis of parameter control of the bottom-level epochs, the number of generations per meta-generation. The $x$-axis represents the number of function evaluations and the $y$-axis representes the meta-fitness, the summed pairwise distance across 10\% of solutions in the map. To compute a single number and indicate variability, Mean $\pm$ SD of the population average meta-fitness is aggregated over 5 replicates.} \label{fig: mr-control}
\end{figure*}

This section provides more data on the development of meta-fitness, the summed pairwise distance across 10\% of solutions in the map. Table~\ref{tab: meta-fitness} displays the final meta-fitness of Meta-evolution algorithms as a function of parameter control strategy. Fig.~\ref{fig: epochs-control} illustrates the effect of controlling the number of generations per meta-generations, with one separate plot for each feature-map (linear, non-linear, and feature-selection). Fig.~\ref{fig: mr-control} analogously illustrates the effect of controlling the mutation rate for different feature-maps.
\begin{table*}[htbp!]
\centering
\caption{Comparison of parameter control methods. The data are divided into 6 groups based on the type of feature-map and the parameter controlled, either the number of generations or the mutation rate. Within each group, bold highlights the condition with highest meta-fitness. Final meta-fitness is is averaged over the final 10\% (i.e., the final 10 million) function evaluations. } \label{tab: meta-fitness}
\subtable[Linear feature-map]
{
\resizebox{!}{0.08\paperheight}{
\begin{tabular}{l l }
\toprule
\textbf{Parameter control} & \textbf{Meta-fitness}   \\ 
\hline
5 generations & $1037.7 \pm 215.9$  \\ 
10 generations & $1037.9 \pm 310.8$  \\ 
25 generations & $952.3 \pm 168.7$  \\ 
50 generations & $718.8 \pm 86.0$  \\ 
Annealing generations & $1036.0 \pm 262.9$  \\ 
RL generations & $\mathbf{1088.7 \pm 373.4}$  \\ 
Endogenous generations & $886.2 \pm 207.6$  \\ 
\hline
Mutation rate 0.125 & $1037.7 \pm 215.9$  \\ 
Mutation rate 0.25 & $1093.7 \pm 499.7$  \\ 
Mutation rate 0.50 & $1185.0 \pm 258.7$  \\ 
Mutation rate 1.0 & $1581.3 \pm 531.4$  \\ 
Annealing mutation rate & $1292.4 \pm 435.6$  \\ 
RL mutation rate & $\mathbf{1640.5 \pm 340.1}$  \\ 
Endogenous mutation rate & $1113.2 \pm 266.4$  \\ 
\bottomrule
\end{tabular}
}
}
\subtable[Non-linear feature-map]{
\resizebox{!}{0.08\paperheight}{
\begin{tabular}{l l }
\toprule
\textbf{Parameter control} & \textbf{Meta-fitness}   \\ 
\hline
5 generations & $14562.2 \pm 1084.8$  \\ 
10 generations & $\mathbf{14716.5 \pm 847.5}$  \\ 
25 generations & $13282.5 \pm 1276.7$  \\ 
50 generations & $12014.3 \pm 1020.3$  \\ 
Annealing generations & $12985.3 \pm 1697.6$  \\ 
RL generations & $13442.5 \pm 868.1$  \\ 
Endogenous generations & $13390.7 \pm 1034.7$  \\ 
\hline
Mutation rate 0.125 & $14562.2 \pm 1084.8$  \\ 
Mutation rate 0.25 & $\mathbf{16470.0 \pm 653.6}$  \\ 
Mutation rate 0.50 & $16327.3 \pm 785.0$  \\ 
Mutation rate 1.0 & $16431.4 \pm 1459.5$  \\ 
Annealing mutation rate & $15292.1 \pm 1198.2$  \\ 
RL mutation rate & $15054.2 \pm 1365.9$  \\ 
Endogenous mutation rate & $14574.3 \pm 1447.8$  \\ 
\bottomrule
\end{tabular}
}
}
\subtable[Selection feature-map]{
\resizebox{!}{0.08\paperheight}{
\begin{tabular}{l l }
\toprule
\textbf{Parameter control} & \textbf{Meta-fitness}   \\ 
\hline
5 generations & $2719.1 \pm 149.0$  \\ 
10 generations & $3009.4 \pm 745.4$  \\ 
25 generations & $2795.7 \pm 515.0$  \\ 
50 generations & $\mathbf{3817.6 \pm 520.2}$  \\ 
Annealing generations & $3356.3 \pm 442.1$  \\ 
RL generations & $3701.7 \pm 662.2$  \\ 
Endogenous generations & $3198.0 \pm 729.0$  \\ 
\hline
Mutation rate 0.125 & $2719.1 \pm 149.0$  \\ 
Mutation rate 0.25 & $4050.7 \pm 350.1$  \\ 
Mutation rate 0.50 & $\mathbf{5067.7 \pm 263.2}$  \\ 
Mutation rate 1.0 & $4755.9 \pm 982.5$  \\ 
Annealing mutation rate & $4896.3 \pm 1019.3$  \\ 
RL mutation rate & $4901.4 \pm 739.8$  \\ 
Endogenous mutation rate & $4781.6 \pm 449.3$  \\ 
\bottomrule
\end{tabular}
}
}
\end{table*}
\newpage
\section{Test comparison of meta-conditions}
\begin{figure}[htbp!]
\centering
\subfigure[Joint 1]{\includegraphics[width=0.30\linewidth]{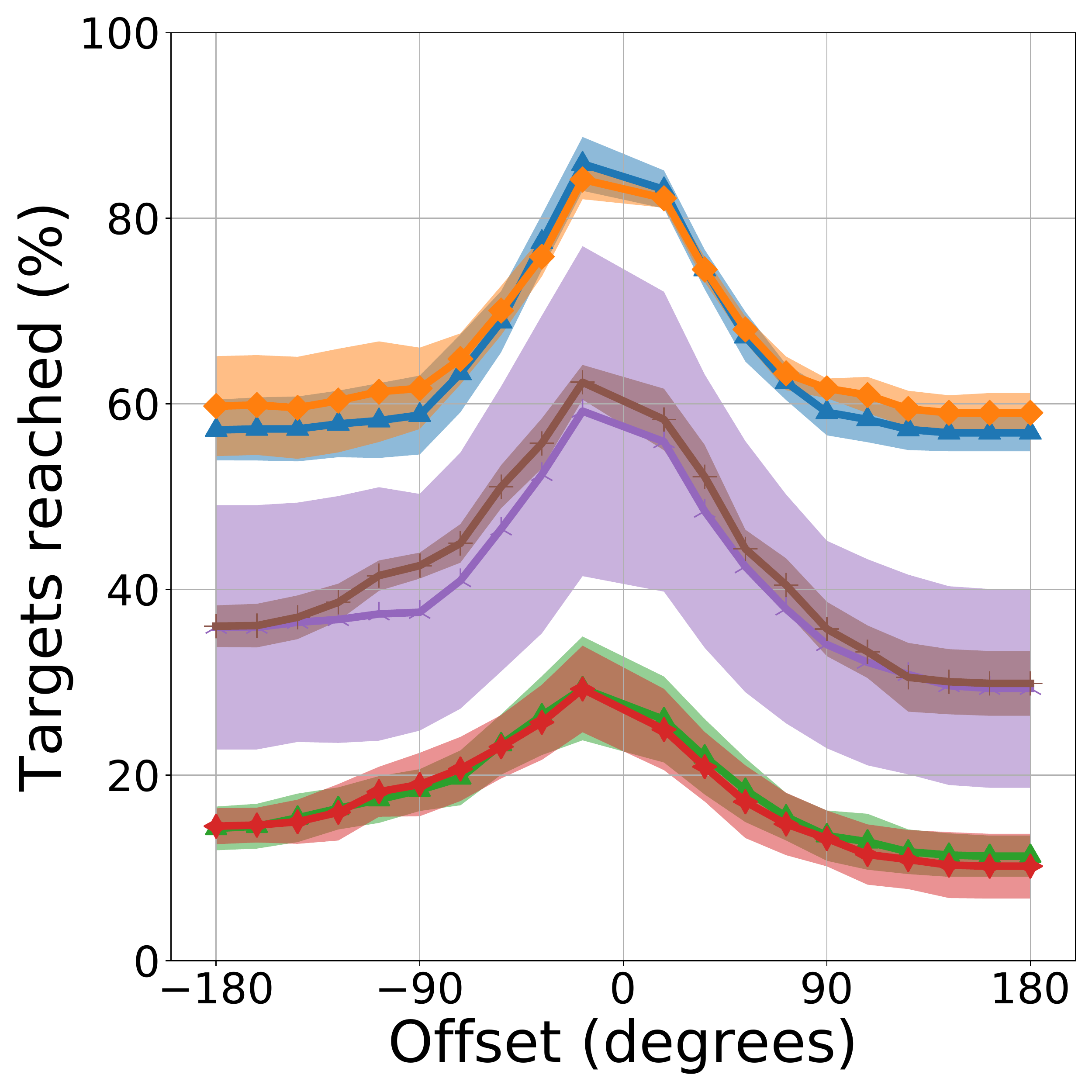}}   
\subfigure[Joint 5]{\includegraphics[width=0.30\linewidth]{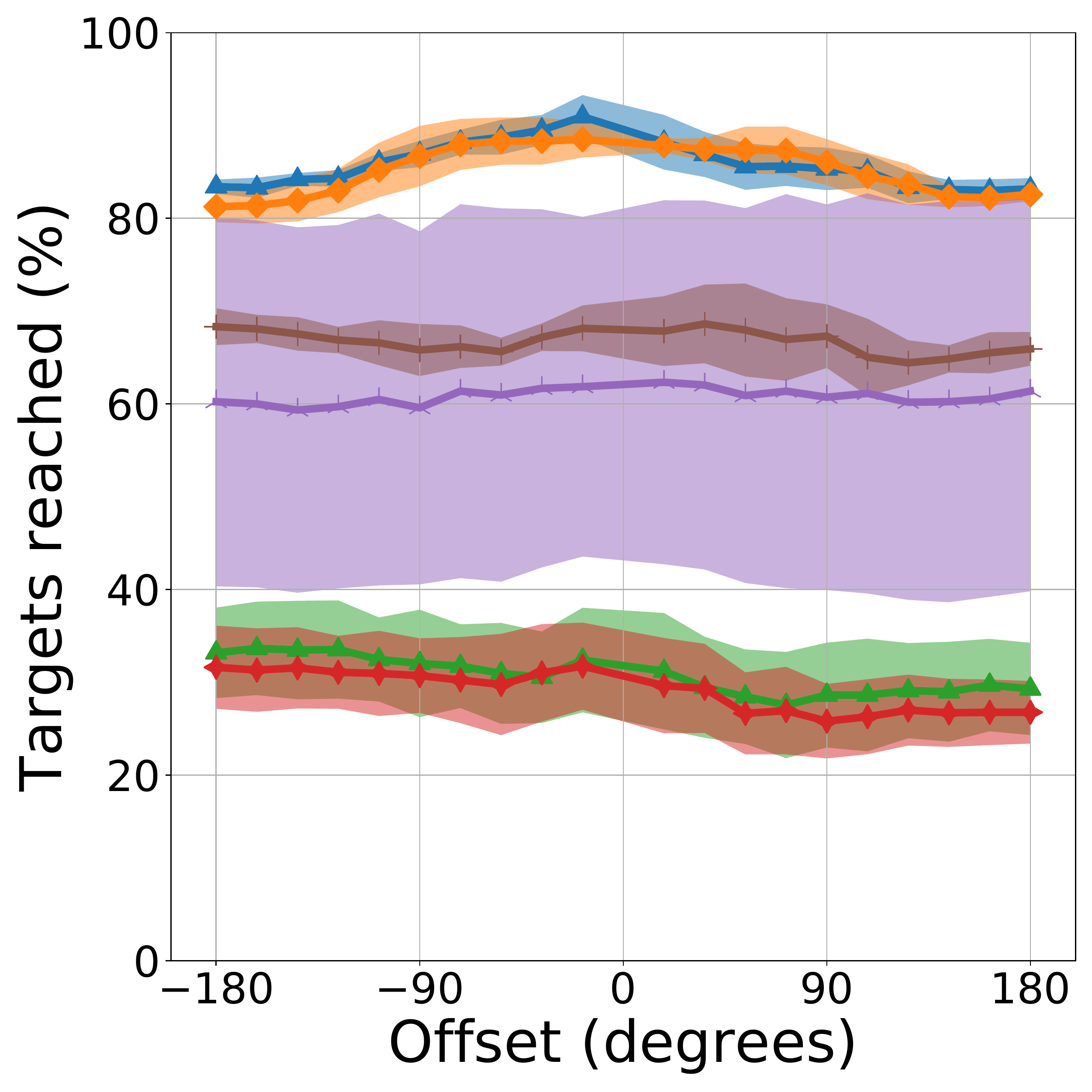}} 
\subfigure[Joint 8]{\includegraphics[width=0.30\linewidth]{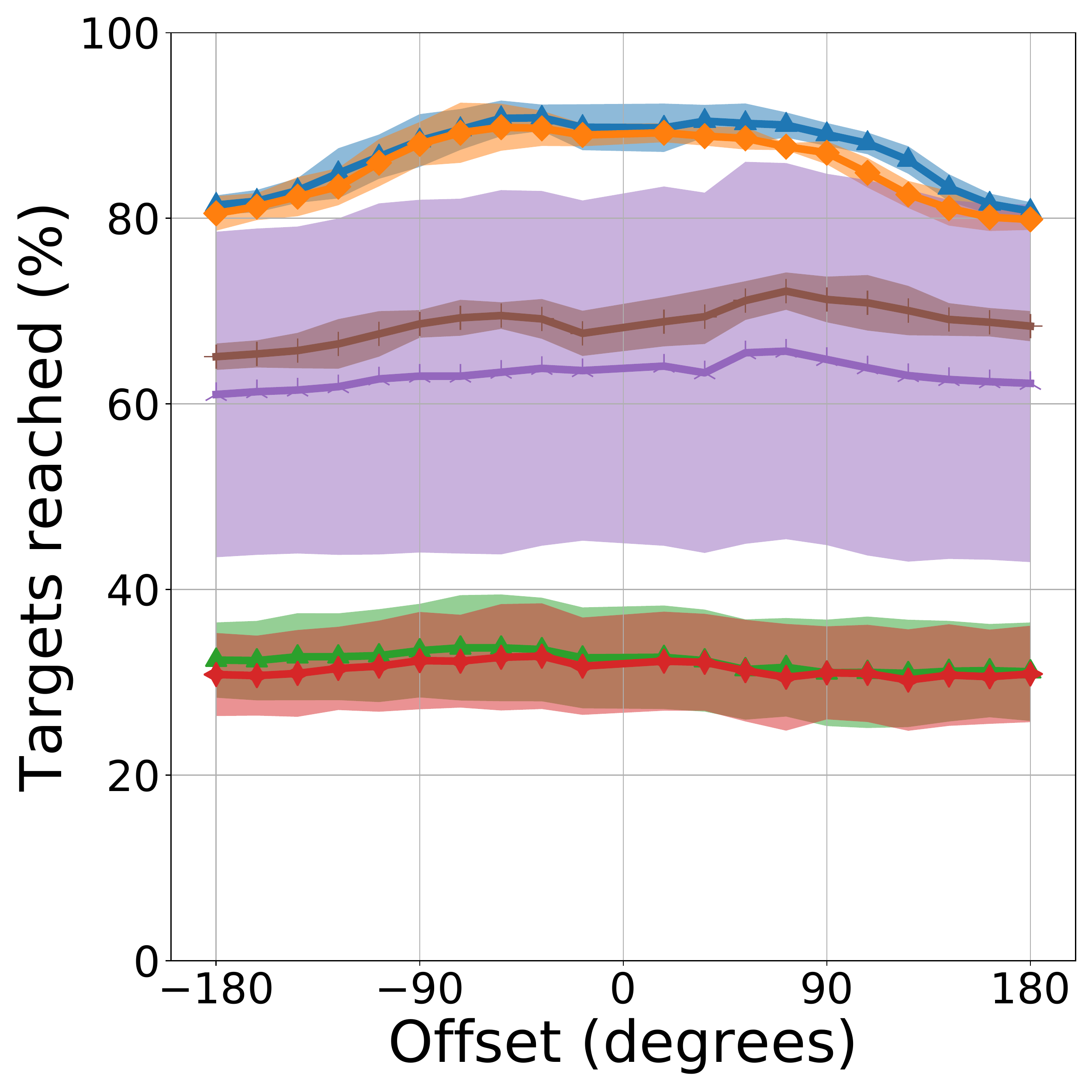}} 
\includegraphics[width=0.90\linewidth]{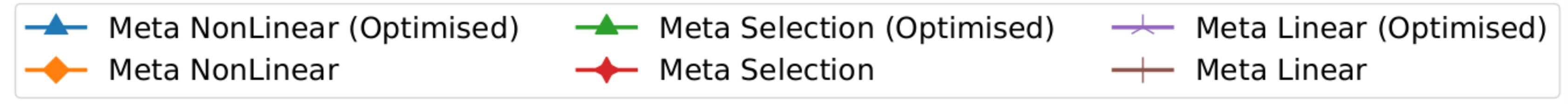}
\caption{Test on unseen damages that offset the joint by a particular angle. The $x$-axis represents the offset in $[-180,180]$ degrees and the $y$-axis represents the percentage of targets reached within the semi-circle span of the robot. For each offset the Mean $\pm$ SD is  aggregated over 5 replicates. The behaviour-performance map is formed from the mean meta-genotype (see $\mathbf{m}$ in Eq.~5). Default parameter settings are a mutation rate of $0.125$ and 5 generations. \textbf{Optimised} indicates the best setting from parameter control: \textbf{RL mutation rate} for Meta Linear, \textbf{Mutation rate 0.25} for Meta NonLinear, and \textbf{Mutation rate 0.50} for Meta Selection. }\label{fig: testperformanceMETA}
\end{figure}

This section provides additional data about the performance of the meta-conditions as they are tested on unseen damages, which offset the desired angle by a particular error in $[-180,180]$ degrees. Table~\ref{tab: significanceMETA} shows summary and significance statistics of damage recovery, with data being aggregated across all offsets. Fig.~\ref{fig: testperformanceMETA} presents the damage recovery for two selected joints as the offset is varied. 
\begin{table}[htbp!]
\centering
\caption{Summary statistics of test on unseen damages. For each meta-condition, we show the percentage of targets reached within the semi-circle span of the robot and, to assess the effect compared to Meta NonLinear (Optimised), the Wilcoxon rank-sum test's significance value and Cliff's delta as an effect size. Bold highlights large effect sizes. The behaviour-performance map is generated from the mean meta-genotype (see $\mathbf{m}$ in Eq.~5). Default parameter settings are a mutation rate of $0.125$ and 5 generations per meta-generation. \textbf{Optimised} indicates the best setting from parameter control: \textbf{RL mutation rate} for Meta Linear, \textbf{Mutation rate 0.25} for Meta NonLinear, and \textbf{Mutation rate 0.50} for Meta Selection.} \label{tab: significanceMETA}
\begin{tabular}{l l l l}
\toprule
\textbf{Condition} & Targets reached ($\%$) & Significance & Cliff's delta  \\ \hline
Meta NonLinear (Optimised) & $79.16 \pm 11.38$ & \quad / & \quad / \\
Meta NonLinear & $79.10 \pm 10.66$  & $p=0.338$ & $0.03$ \\ 
Meta Selection (Optimised) & $26.86 \pm 8.02$ & $p<0.001$ & $\mathbf{1.00}$ \\ 
Meta Selection & $25.82 \pm 7.80$ & $p<0.001$ & $\mathbf{1.00}$\\ 
Meta Linear (Optimised) & $54.85 \pm 21.13$ & $p<0.001$ & $\mathbf{0.72}$ \\ 
Meta Linear & $59.35 \pm 12.50$ & $p<0.001$ & $\mathbf{0.74}$ \\ 
\bottomrule
\end{tabular}
\end{table}

\section{Source code}
Source code for the experiments is publicly available at \url{https://github.com/resilient-swarms/planar_metacmaes}.